\documentclass{article}

\PassOptionsToPackage{numbers,sort&compress}{natbib}
\usepackage[preprint]{neurips_2026}

\usepackage[utf8]{inputenc}
\usepackage[T1]{fontenc}
\usepackage{hyperref}
\usepackage{url}
\usepackage{booktabs}
\usepackage{amsfonts}
\usepackage{amsmath}
\usepackage{amssymb}
\usepackage{nicefrac}
\usepackage{microtype}
\usepackage{xcolor}
\usepackage{graphicx}
\usepackage{enumitem}
\usepackage{algorithm}
\usepackage{algorithmic}
\usepackage{pifont}
\usepackage{multirow}
\usepackage{wrapfig}
\usepackage{subcaption}
\usepackage{adjustbox}
\usepackage{placeins}
\usepackage{tikz}
\usetikzlibrary{arrows.meta, positioning, shapes.geometric, fit, backgrounds}

\graphicspath{{./}}

\title{Beyond Text Prompts: Visual-to-Visual Generation as A Unified Paradigm}

\author{%
  \textbf{Yaofang Liu$^{1}\textsuperscript{\dag}$} \quad
  \textbf{Kangning Cui$^{2}$} \quad
  \textbf{Meng Chu$^{3}$} \quad
  \textbf{Zhaoqing Li$^{4}$} \quad \\
  \textbf{Suiyun Zhang$^{5}$} \quad
  \textbf{Jean-Michel Morel$^{7}$} \quad
  \textbf{Xiaodong Cun$^{6}$} \quad
  \textbf{Haoxuan Che$^{5}\textsuperscript{\dag}\textsuperscript{*}$} \quad \\
  \textbf{Rui Liu$^{5}\textsuperscript{\dag}$} \quad
  \textbf{Raymond H. Chan$^{7}$} \\
  \\
  $^{1}$City University of Hong Kong \quad
  $^{2}$City University of Hong Kong (Dongguan) \\
  $^{3}$The Hong Kong University of Science and Technology \quad
  $^{4}$The Chinese University of Hong Kong \\
  $^{5}$Celia Research HK \quad
  $^{6}$Great Bay University \quad
  $^{7}$Lingnan University \\
  \textsuperscript{*}Project lead \quad
  \textsuperscript{\dag}Corresponding authors \\
  {\textbf{Project Page:
  \url{https://yaofang-liu.github.io/V2V_Web/}
  }}
}

\begin{document}

\maketitle

\begin{abstract}
Humans often understand, specify, and create through visual artifacts: typography sheets, sketches, reference images, and annotated scenes. Yet modern visual generators still usually ask users to serialize this intent into text. This text-first interface is convenient, but it is also a bottleneck: natural language greatly compresses visual signals like spatial structure, exact appearance, and glyph shape. We propose \textbf{\emph{visual-to-visual} (V2V)} generation, in which the user conditions a generative model with a visual specification page rather than a text prompt. The page is not an edit target to reconstruct, but a visual document that specifies the desired output. We introduce \textbf{V2V-Zero}, a training-free framework that exposes this interface in existing vision-language model (VLM) conditioned generators by replacing text-only user conditioning with final-layer hidden states extracted from visual pages. The key observation is architectural: when a diffusion generator is trained to consume hidden states from a VLM, the frozen VLM already maps both text and images into the conditioning space used by the generator.

On GenEval, V2V-Zero achieves 0.85 overall with a frozen Qwen-Image backbone, closely matching the backbone's optimized text-to-image (T2I) performance without fine-tuning. To evaluate the broader V2V space, we introduce \textbf{Simple-V2V Bench}, a novel benchmark spanning seven visual-conditioning tasks and seven evaluated models, including GPT Image 2, Nano Banana 2, Seedream 5.0 Lite, open-weight image models, and a video extension. V2V-Zero scores 32.7/100 overall, outperforming evaluated open-weight image baselines and revealing a consistent capability hierarchy: visual attribute binding is already strong, content generation remains unreliable, and structural control remains difficult even for evaluated commercial systems. A HunyuanVideo-1.5 extension scores 20.2/100, providing complementary evidence that the same visual-page interface transfers beyond still images. Mechanistic analysis shows that the default reasoning path is primarily visually routed: real DiT attention assigns 95.0\% of conditioning-token mass to visual-page hidden states rather than generated reasoning-token states. Together, these results identify V2V as both an immediately available zero-shot interface and a research direction for native generators trained to \textbf{treat visual input as a first-class conditioning language.}

\end{abstract}

% \clearpage

\begin{figure}[t]
  \centering
  \captionsetup{skip=3pt}
  \includegraphics[width=1\linewidth,height=0.7\textheight,keepaspectratio]{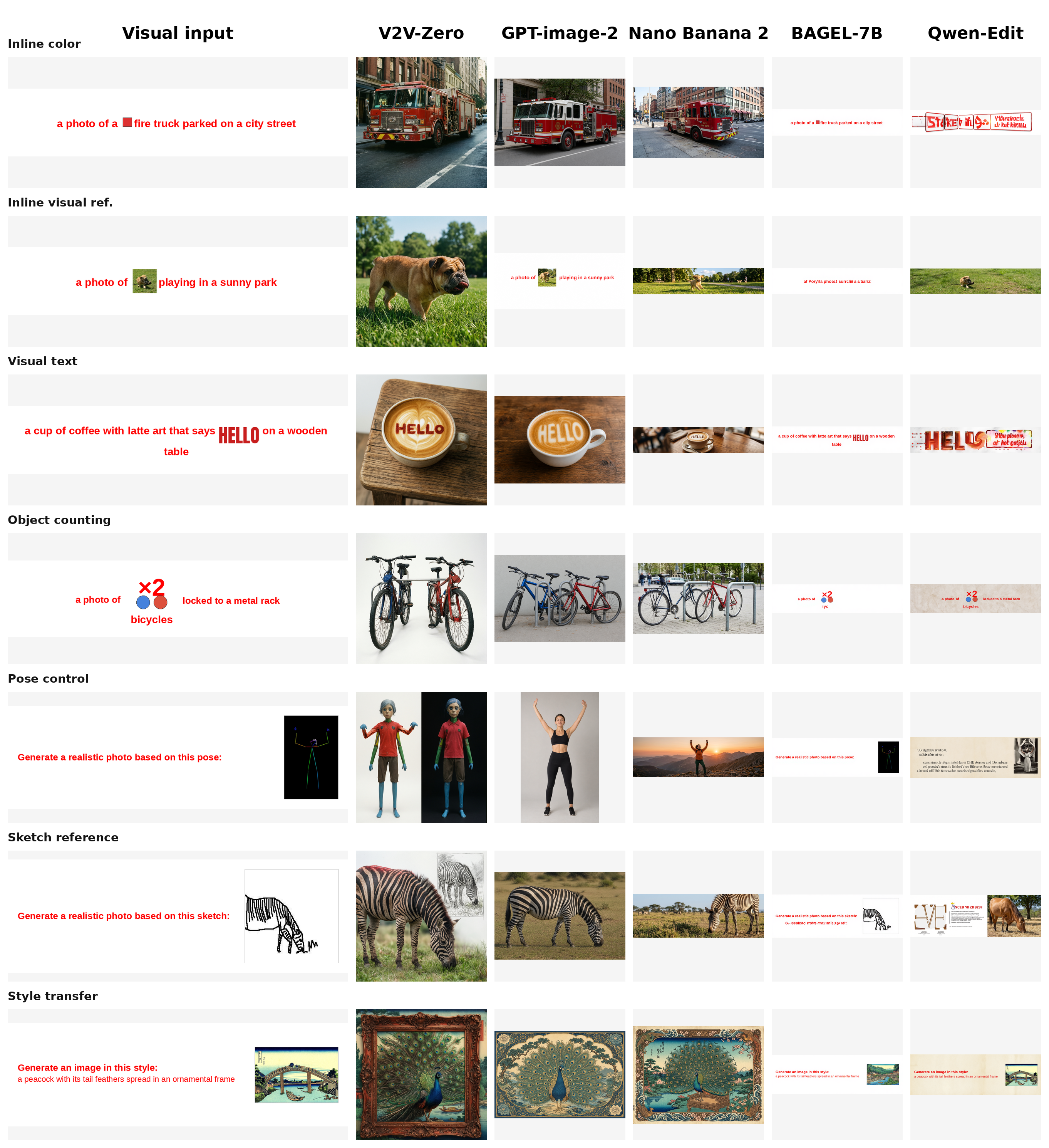}
  \caption{\textbf{Qualitative Simple-V2V Bench comparison.} Rows are visual-conditioning tasks and columns compare the same visual page across V2V-Zero and SOTA baselines, previewing strong attribute/reference binding and harder counting, pose, sketch, and style-transfer cases.}
  \label{fig:v2vbench_qualitative}
  \vspace{-0.5em}
\end{figure}

%==============================================================================
% SECTION 1: INTRODUCTION
%==============================================================================

\section{Introduction}
\label{sec:intro}

Human visual intent is rarely born as a sentence. Designers use sketches, palettes, reference boards, typography sheets, pose diagrams, spatial layouts, and annotated images; ordinary users point to examples, colors, arrangements, and glyphs. These artifacts are not auxiliary explanations of a text prompt. They are often the most direct representation of what should be generated. Nevertheless, the dominant interface to modern image and video generators remains text: the user must translate visual intent into a sequence of discrete tokens, and the model must reconstruct the intended visual constraints from that serialization.

This text-first convention is increasingly misaligned with both human communication and model architecture. Spatial relations, counts, color bindings, subject appearance, typography, and geometric structure are naturally visual. A blueprint specifies placement more directly than a paragraph; a swatch specifies color more precisely than a color name; rendered glyphs specify character shape and style without requiring the model to infer them from language; an inline thumbnail can bind an object reference at the exact position where a noun would otherwise appear. Text remains powerful, but visual broadens its role: any linguistic instruction can also be rendered as visual content, with fonts, styles, colors, layout, and surrounding references that make the specification more diverse and expressive than a text prompt alone.

We therefore propose \emph{visual-to-visual} (V2V) generation as a paradigm in which the user-facing conditioning input is a structured visual specification page. A V2V page may contain rendered text, sketches, color chips, reference images, spatial diagrams, style examples, or other visual evidence that defines the target output. This is distinct from image editing: the input page is not an image to be reconstructed or modified, but a conditioning document whose visual content should be interpreted as the scene specification. V2V also differs from task-specific control modules, because the same page interface can combine heterogeneous visual cues in a single document.

The reason this paradigm is immediately testable is that modern text-to-visual systems increasingly expose a latent route from multimodal understanding to generation. In architectures such as Qwen-Image \cite{wu2025qwen}, a multimodal VLM conditioning encoder reads the prompt context, and the diffusion transformer cross-attends to the resulting final-layer hidden states~\cite{bai2025qwen25vltechnicalreport}. If the conditioning encoder natively accepts both text and images, then visual inputs can be mapped into the same hidden-state space already consumed by the generator. This does not require changing model weights, adding an adapter, or training a new controller. It changes the user-side conditioning variable: instead of feeding scene content as text, we feed a visual page through the frozen VLM and expose its hidden states to the generator's existing conditioning interface.

% This observation defines the scope of our claim. V2V is not a single benchmark task, nor a claim that visual pages solve all forms of control zero-shot. It is a general conditioning interface for VLM-based image and video generators. Its effectiveness depends on whether the frozen VLM encodes the page into generation-compatible hidden states and whether the frozen diffusion model has learned to use those states. We analyze this mechanism in Section~\ref{sec:mechanism}, verify it primarily through T2I experiments, and use HunyuanVideo-1.5 as complementary evidence that the same framework extends to text-to-video (T2V) generation.

\begin{figure}[t]
\centering
\includegraphics[width=\linewidth]{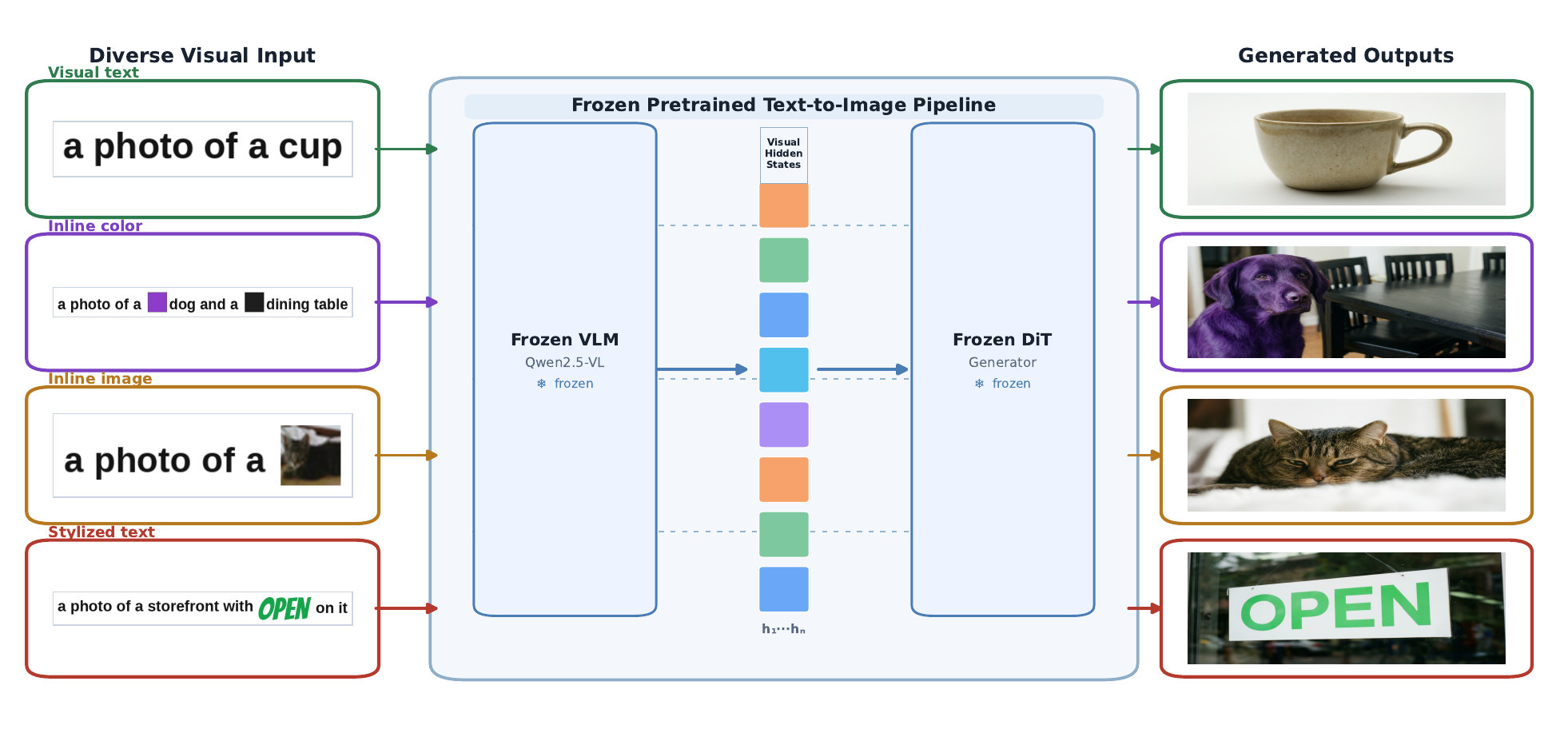}
\caption{\textbf{V2V-Zero replaces user text prompts with visual prompt pages.} A frozen VLM can accept plain visual text, inline color blocks, inline image blocks, or stylized rendered text tokens as encoder inputs. The main V2V-Zero path keeps pretrained weights and learned modules unchanged: the VLM reads the visual page, exposes visual hidden states, and the frozen DiT generator cross-attends to those states through its existing conditioning interface.}
\label{fig:implicit_v2v_insight}
\end{figure}

Towards this, we then introduce \textbf{V2V-Zero}, a systematic zero-shot instantiation of this paradigm. V2V-Zero builds visual specification pages, processes them with a frozen VLM encoder, and conditions a frozen diffusion model by direct final-layer hidden-state injection. We use two conditioning modes: Image-HS-only as a non-reasoning visual-state control, and \textsc{Full-Final} as the default reasoning path. Figure~\ref{fig:v2vbench_qualitative} gives the qualitative preview of the benchmark comparison, while Figure~\ref{fig:implicit_v2v_insight} illustrates the core mechanism: the generator's learned conditioning interface is reused, but the frozen VLM receives user-provided visual evidence rather than only a linguistic prompt. On the GenEval benchmark~\cite{ghosh2023geneval}, V2V-Zero achieves 0.85 overall, closely comparable to the backbone's own optimized T2I performance. To evaluate the broader V2V space, we introduce Simple-V2V Bench, a novel benchmark across seven visual-conditioning tasks evaluated with seven models including GPT Image 2, Nano Banana 2, and Seedream 5.0 Lite. The benchmark reveals a three-tier capability structure: attribute binding is already strong, content generation remains unreliable, and structural control such as pose and sketch following remains difficult even for the strongest evaluated systems. Finally, applying the same V2V-Zero method to HunyuanVideo-1.5~\cite{wu2025hunyuanvideo} scores 20.2/100, providing complementary evidence that the framework applies to T2V as well as T2I.

\noindent\textbf{Contributions.}
\begin{itemize}
    \item We formulate \textbf{V2V as a unified image and video generation paradigm} and present \textbf{V2V-Zero}, a training-free framework that processes visual pages through a frozen VLM and injects its hidden states into a frozen diffusion model. On GenEval, V2V-Zero achieves \textbf{0.85 overall}, comparable to the backbone's own optimized T2I performance without any finetuning.
    \item We introduce \textbf{Simple-V2V Bench}, a simple and novel benchmark across \textbf{seven visual-conditioning tasks} and \textbf{seven models} that maps the capability frontier: attribute binding is strong, content generation remains unreliable, and structural control is still difficult.
    \item We provide a \textbf{T2V validation} by applying the same visual-page conditioning path to HunyuanVideo-1.5, scoring \textbf{20.2/100} and demonstrating that the V2V framework extends to text-to-video generation.
\end{itemize}

\section{Related work}
\label{sec:related}

T2I and T2V models have advanced rapidly through diffusion, transformer, and multimodal-conditioning backbones~\cite{ddpm,ldm,imagen,sdxl,peebles2023scalable,liu2024redefining,wu2025qwen,makeavideo,yang2024cogvideox,liupusa,polyak2024movie,chen2023videocrafter1,wu2025hunyuanvideo,liu2024evalcrafter}, while editing systems typically treat images or videos as sources to modify under textual, inversion, attention-control, instruction-tuning, or temporal-propagation mechanisms~\cite{meng2021sdedit,hertz2022prompt,brooks2023instructpix2pix,geyer2023tokenflow,ku2024anyv2v}. A second line adds task-specific visual controls: structural maps, boxes, reference images, sketches, motion, all-in-one control interfaces, and glyph-aware text rendering~\cite{zhang2023adding,mou2024t2i,li2023gligen,ye2023ip,ruiz2023dreambooth,blattmann2023stable,wang2023videocomposer,xiao2025omnigen,wu2026omnigen2instructionalignedmultimodalgeneration,jiang2025vace,liu2023character,yang2023glyphcontrol,tuo2023anytext,chen2023textdiffuser,betker2023improving}. The closest conceptual work studies images as prompts, context, demonstrations, or native multimodal tokens~\cite{bar2022visual,wang2023context,najdenkoska2024context,oorloff2025stable,li2025visualcloze,lin2025realgeneral,ye2025unic,cui2025emu3,xie2025show,deng2025emerging,liu2026tuna,wang2026deepgen}. V2V differs by treating a single visual specification page as the user interface itself: text can be rendered into vision, and color swatches, glyphs, thumbnails, sketches, style references, layouts, and temporal cues can coexist without adding a task-specific adapter. A fuller discussion is provided in Appendix~\ref{app:related_work}.

%==============================================================================
% SECTION 3: METHOD
%==============================================================================

\section{Method}
\label{sec:method}

V2V-Zero instantiates the V2V paradigm by conditioning a frozen diffusion generator on structured visual specification pages processed by a frozen VLM. The key insight is that user-provided visual inputs---spatial blueprints, color cards, rendered text, inline thumbnails---carry scene information that text prompts often compress, and the VLM already maps those inputs into the generator's conditioning space. In the primary T2I setting, the generator cross-attends directly to final-layer VLM visual hidden states. Figure~\ref{fig:implicit_v2v_insight} illustrates this: visual text, inline color blocks, inline image blocks, and stylized rendered text are different encoder inputs for the same unchanged VLM-to-DiT conditioning path. Because recent T2I and T2V systems have converged toward the same VLM-hidden-state-to-diffusion architecture, the same abstraction also applies to a VLM-conditioned video generator: the visual page is encoded by the model's multimodal conditioning encoder, then the resulting hidden states replace or augment the standard text prompt embeddings before temporal denoising.

\subsection{Problem formulation}
\label{sec:formulation}

Let $V \in \mathbb{R}^{H \times W \times 3}$ be the user-provided visual specification page that encodes the target scene through layout, swatches, thumbnails, or rendered text. Let $\mathcal{E}$ denote the frozen multimodal VLM conditioning encoder, distinct from the visual input variable $V$, and let $\mathcal{G}$ denote the frozen diffusion generator. V2V-Zero produces
\begin{equation}
  I = \mathcal{G}\!\bigl(\mathcal{E}(V)\bigr),
  \label{eq:pipeline}
\end{equation}
where $\mathcal{E}(V)$ denotes the final VLM hidden states consumed directly by the generator. The equation makes the user-side paradigm shift explicit: the conditioning variable is the visual page $V$, which replaces the text user input, and the VLM output itself is the generator conditioning. Fixed templates, system instructions, and prompt wrappers remain ordinary VLM/T2I scaffolding for interpreting the page; they are not the user-controlled input in Eq.~\ref{eq:pipeline} and do not supply GenEval scene content. Implementation details and pseudocode are deferred to Appendix~\ref{app:implementation} and Appendix~\ref{app:algorithm}.

V2V-Zero is strictly training-free: it performs no weight updates, trains no adapter, and adds no learned module. The only intervention is an inference-time conditioning wrapper that exposes final-layer VLM hidden states and feeds them to the generator's prompt-conditioning slot. The visual page $V$ is processed only by the VLM vision encoder as conditioning input. In the strongest compositional setting, $V$ is an inline visual prompt page whose local swatches or thumbnails are embedded directly in a single rendered prompt line. For the HunyuanVideo-1.5~\cite{wu2025hunyuanvideo} validation, we use the same principle with the video pipeline's multimodal VLM encoder: a rendered Simple-V2V Bench page is encoded as visual prompt context, padded or truncated to the model's conditioning length, and then used for full T2V sampling without training. This experiment tests the T2V direction of the same framework; the most extensive benchmark evidence remains image-centered.

\subsection{Conditioning modes}
\label{sec:extraction}

V2V-Zero uses two conditioning modes. \emph{Image-HS-only} is the non-reasoning control: the generator cross-attends only to image-token states extracted from the visual page,
\begin{equation}
  \mathcal{E}_{\textsc{img}}(V) = [H_{\mathrm{img}}].
\end{equation}
\textsc{Full-Final} is the default reasoning path. Let the fixed prefix be
\begin{equation}
  \mathbf{x}=[t_{\mathrm{sys}}, \operatorname{ViT}(V), t_{\mathrm{user}}(T), \langle\mathrm{gen}\rangle].
\end{equation}
The VLM autoregressively generates reasoning tokens $t_{1:N}$ from $\mathbf{x}$. We then recompute the final-layer hidden states under teacher forcing,
\begin{equation}
  \widetilde{H}^{(L)}=\mathcal{E}^{(L)}([\mathbf{x},t_{1:N}]),
\end{equation}
and inject
\begin{equation}
  H_{\textsc{full\text{-}final}} =
  [\widetilde{H}^{(L)}_{\operatorname{ViT}(V)}; \widetilde{H}^{(L)}_{t_{1:N}}],
\end{equation}
namely the visual states plus the generated-token final states, where $L$ denotes the final VLM layer index. Recomputing means running one forward pass on the fixed prefix and generated tokens so all injected states come from the same final-layer context. We use \textsc{Full-Final} by default because it preserves direct visual states while adding VLM reasoning about the page. The generated reasoning states should be understood as part of the conditioning sequence, not as a decoded-text replacement for the visual page; Section~\ref{sec:mechanism} shows that the DiT predominantly attends to the visual-prefix states in this path. For Qwen-Image, all direct injection uses the final VLM layer ($L=28$); Appendix~\ref{sec:layerwise} shows that earlier layers fail. Note that this last-layer setting is specifically for Qwen-Image, for HunyuanVideo-1.5 extension, we follow its default setting to use the third-from-last layer (Section~\ref{sec:hunyuan_video}).

\subsection{Templates and visual pages}
\label{sec:templates}
\label{sec:pages}

Templates are fixed interpreter instructions that specify how the VLM should read the visual page; they do not supply scene content. We demonstrate three visual-page families: \emph{compositional pages} (spatial diagrams for structured scene control), \emph{text pages} (rendered target characters), and \emph{inline visual pages} (swatches and thumbnails embedded within prompt text). The full eight-page taxonomy, per-family construction details, and exact template text are provided in Appendix~\ref{app:page_taxonomy}.

%==============================================================================
% SECTION 4: EXPERIMENTS
%==============================================================================

% =============================================================================
% Experiments
% =============================================================================

\section{Experiments}
\label{sec:experiments}

\subsection{Setup}
\label{sec:setup}

We evaluate primarily on image generation, where established benchmarks and baselines allow controlled measurement. GenEval~\cite{ghosh2023geneval} contains 553 prompts over six compositional skills, with 4 samples per prompt. Our generator follows Qwen-Image~\cite{wu2025qwen} with Qwen2.5-VL-7B-Instruct as the conditioning backbone~\cite{wang2024qwen2,bai2025qwen25vltechnicalreport}. For V2V-Zero, the original text prompt is rendered as a visual page; the generator receives VLM hidden states derived from that page. Unless noted otherwise, decoding is greedy and scores follow the benchmark's standard aggregation. We also introduce Simple-V2V Bench (Section~\ref{sec:v2v_bench}) and validate the same path on T2V generation with HunyuanVideo-1.5 (Section~\ref{sec:hunyuan_video}).

\subsection{Main results}
\label{sec:main_results}

Table~\ref{tab:main_results} compares V2V-Zero against T2I models on the full GenEval benchmark. With the same Qwen-Image-2512 backbone, \textsc{Full-Final} reaches 0.85 without fine-tuning or additional modules. It approaches the official optimized Qwen-Image result while using a visual conditioning route with entirely frozen weights.

\begin{table}[t]
  \centering
  \caption{Quantitative evaluation on GenEval~\cite{ghosh2023geneval} (553 prompts, 4 samples each). Best per-column in bold.}
  \label{tab:main_results}
  \small
  \begin{adjustbox}{width=\textwidth}
  \begin{tabular}{l|cccccc|c}
    \toprule
    \multirow{2}{*}{\textbf{Model}} & \textbf{Single} & \textbf{Two} & \multirow{2}{*}{\textbf{Counting}} & \multirow{2}{*}{\textbf{Colors}} & \multirow{2}{*}{\textbf{Position}} & \textbf{Attribute} & \multirow{2}{*}{\textbf{Overall$\uparrow$}} \\
    & \textbf{Object} & \textbf{Object} & & & & \textbf{Binding} & \\
    \midrule
    Show-o~\cite{xie2024show} & 0.95 & 0.52 & 0.49 & 0.82 & 0.11 & 0.28 & 0.53 \\
    Emu3-Gen~\cite{wang2024emu3} & 0.98 & 0.71 & 0.34 & 0.81 & 0.17 & 0.21 & 0.54 \\
    PixArt-$\alpha$~\cite{chen2023pixart} & 0.98 & 0.50 & 0.44 & 0.80 & 0.08 & 0.07 & 0.48 \\
    SD3 Medium~\cite{esser2024scaling} & 0.98 & 0.74 & 0.63 & 0.67 & 0.34 & 0.36 & 0.62 \\
    FLUX.1 [Dev]~\cite{labs2025flux1kontextflowmatching} & 0.98 & 0.81 & 0.74 & 0.79 & 0.22 & 0.45 & 0.66 \\
    SD3.5 Large~\cite{esser2024scaling} & 0.98 & 0.89 & 0.73 & 0.83 & 0.34 & 0.47 & 0.71 \\
    JanusFlow~\cite{ma2025janusflow} & 0.97 & 0.59 & 0.45 & 0.83 & 0.53 & 0.42 & 0.63 \\
    Lumina-Image 2.0~\cite{qin2025lumina} & -- & 0.87 & 0.67 & -- & -- & 0.62 & 0.73 \\
    Janus-Pro-7B~\cite{chen2025janus} & 0.99 & 0.89 & 0.59 & 0.90 & \textbf{0.79} & 0.66 & 0.80 \\
    HiDream-I1-Full~\cite{cai2025hidream} & \textbf{1.00} & \textbf{0.98} & 0.79 & 0.91 & 0.60 & 0.72 & 0.83 \\
    GPT Image 1~\cite{openai_gpt_image_1} & 0.99 & 0.92 & 0.85 & 0.92 & 0.75 & 0.61 & 0.84 \\
    Seedream 3.0~\cite{gao2025seedream} & 0.99 & 0.96 & \textbf{0.91} & \textbf{0.93} & 0.47 & \textbf{0.80} & 0.84 \\
    \midrule
    Qwen-Image (official)~\cite{wu2025qwen} & 0.99 & 0.92 & 0.89 & 0.88 & 0.76 & 0.77 & \textbf{0.87} \\
    Qwen-Image (reproduced) & 0.99 & 0.97 & 0.87 & 0.89 & 0.71 & 0.74 & 0.86 \\
    % Qwen-Image-2512 (text-only, no magic) & 0.98 & 0.92 & 0.68 & 0.87 & 0.49 & 0.56 & 0.75 \\
    % \midrule
    % \textbf{V2V-Zero: Image HS only}$^\dagger$ & -- & -- & -- & -- & -- & -- & 0.72 \\
    \textbf{Qwen-Image-V2V-Zero: Full-final} & \textbf{1.00} & 0.95 & 0.85 & 0.89 & 0.73 & 0.68 & 0.85 \\
    \bottomrule
  \end{tabular}
  \end{adjustbox}
\end{table}

Image-HS-only achieves 71.57\% on the ablation subset; \textsc{Full-Final} adds VLM reasoning-token states and reaches 86.77\% on the same subset, making it the default mode. More results are reported in Appendix~\ref{app:full_ablation}.Section~\ref{sec:mechanism} measures how the generator uses this path: real DiT attention assigns 95.0\% of conditioning-token mass to the visual tokens' states.

\subsection{Ablations}
\label{sec:ablation}

Full ablation results are tabulated in Appendix~\ref{app:full_ablation}; this section summarizes the key findings.

\subsubsection{Conditioning mode ablation}
\label{sec:ablation_mode}

We use two conditioning modes. Image-HS-only isolates the visual hidden states without autoregressive reasoning. \textsc{Full-Final} keeps those image states and appends final-layer reasoning states, improving the diagnostic-subset score from 71.57\% to 86.77\% (+15.20\,pp). We therefore use \textsc{Full-Final} as the default mode for GenEval, Simple-V2V Bench, and the video extension; the mechanism analysis shows that this default remains primarily visually routed inside the DiT.

\subsubsection{Other ablations}
\label{sec:ablation_tokens}
\label{sec:ablation_template}
\label{sec:ablation_other}

Token count exhibits a non-monotonic sweep; the strongest fullbench setting uses 200 tokens. Template choice affects how the VLM interprets the visual page, as expected for VLM-conditioned systems: a generation-oriented template at 100 tokens reaches 84.56\%, while a generic description template drops to 62.01\%. The fixed template does not supply GenEval scene content. Using the final VLM layer is essential (Appendix~\ref{sec:layerwise}).

\subsection{Simple-V2V Bench}
\label{sec:v2v_bench}

The V2V paradigm opens a generation space broader than conventional T2I or image editing. We introduce \textbf{Simple-V2V Bench}, a seven-category benchmark (22 prompts each) where the model must interpret a user-provided visual specification page and generate the corresponding image.

\noindent\textbf{Benchmark design.}
Each prompt pairs a visual input page with a generation target. The categories cover inline color, inline visual reference, visual text, style transfer, object counting, sketch reference, and pose control; full construction details are in Appendix~\ref{app:simple_v2v_construction}.

\noindent\textbf{Evaluation.}
We use Qwen3-VL-32B-Instruct as a direct VLM judge: it sees both the input page and generated output, scores \emph{Quality} and \emph{Alignment} on 1--10 scales, and reports $\min(\text{Quality}, \text{Alignment}) \times 10$. Scores are averaged over 4 samples per prompt; Appendix~\ref{app:simple_v2v_protocol} gives the rubrics and deduction rules.

\noindent\textbf{Results and analysis.}
Table~\ref{tab:v2v_bench_baselines} reports V2V-Zero's performance alongside six baselines evaluated under the same protocol (detailed V2V-Zero per-category scores including Quality and Alignment breakdown in Appendix~\ref{app:simple_v2v_results}). V2V-Zero scores 32.7/100 overall, revealing a clear capability gradient across task types. Among baselines, GPT Image 2~\cite{openai_gpt_image_2} leads at 64.7, followed by Seedream 5.0 Lite~\cite{seedream_5_lite} (58.4) and Nano Banana 2~\cite{nano_banana2} (51.4), while open-weight models score substantially lower: Qwen-Image-Edit-2511~\cite{wu2025qwen} (19.7) and BAGEL-7B-MoT~\cite{deng2025emerging} (15.2). All Simple-V2V scores are from our own evaluation. HunyuanVideo-1.5 generates full videos from the same visual pages and scores 20.2 when those videos are evaluated by the same VLM judge on the generated video.

% MOVED TO APPENDIX per revision
%\begin{table}[t]
%  \centering
%  \caption{\textbf{Simple-V2V Bench} results for V2V-Zero (scored by Qwen3-VL-32B, 4 samples per prompt). Each sample receives independent Quality (Q) and Alignment (A) scores on a 1--10 scale; the final score is $\min(\text{Q},\text{A})\times 10$. Categories ordered by V2V-Zero score.}
%  \label{tab:v2v_bench}
%  \small
%  \setlength{\tabcolsep}{4pt}
%  \begin{tabular}{lcccc}
%    \toprule
%    \textbf{Category} & \textbf{Score} & \textbf{Quality} & \textbf{Alignment} & \textbf{Insight} \\
%    \midrule
%    Inline color & 76.9 & 9.51 & 7.69 & Semantic binding works \\
%    Inline visual ref & 42.8 & 8.01 & 4.28 & Object identity partially preserved \\
%    Visual text & 34.8 & 6.58 & 3.48 & Spelling errors dominate \\
%    Object counting & 24.0 & 5.39 & 2.40 & Count frequently wrong \\
%    Style transfer & 20.3 & 7.58 & 2.03 & Style often ignored \\
%    Sketch reference & 16.6 & 5.44 & 1.66 & Layout not followed \\
%    Pose control & 13.3 & 2.98 & 1.33 & Structural failure \\
%    \midrule
%    \textbf{Overall} & \textbf{32.7} & 6.50 & 3.27 & \\
%    \bottomrule
%  \end{tabular}
%\end{table}
% MOVED TO APPENDIX per revision

\begin{table}[t]
  \centering
  \caption{\textbf{Simple-V2V Bench: multi-model comparison.} All image models are evaluated with the same Qwen3-VL-32B judge, dual-dimension scoring ($\min(\text{Q},\text{A})\times 10$), and mean-of-4-samples aggregation. HunyuanVideo generates full videos; the reported score applies the same VLM judge directly to the generated video. Best per-category in bold.}
  \label{tab:v2v_bench_baselines}
  \small
  \setlength{\tabcolsep}{3pt}
  \begin{adjustbox}{width=\textwidth}
  \begin{tabular}{l|ccccccc|c}
    \toprule
    \textbf{Model} & \textbf{Vis.Text} & \textbf{Inl.Color} & \textbf{Inl.VisRef} & \textbf{Counting} & \textbf{Style} & \textbf{Pose} & \textbf{Sketch} & \textbf{Overall} \\
    \midrule
    GPT Image 2~\cite{openai_gpt_image_2} & 78.3 & \textbf{92.4} & 75.8 & \textbf{91.8} & \textbf{60.3} & \textbf{20.0} & \textbf{34.0} & \textbf{64.7} \\
    Seedream 5.0 Lite~\cite{seedream_5_lite} & \textbf{79.0} & 68.7 & 74.7 & 88.8 & 48.7 & 16.8 & 32.4 & 58.4 \\
    Nano Banana 2~\cite{nano_banana2} & 59.2 & 69.7 & \textbf{78.0} & 67.1 & 44.7 & 19.1 & 22.3 & 51.4 \\
    \midrule
    V2V-Zero (ours) & 34.8 & 76.9 & 42.8 & 24.0 & 20.3 & 13.3 & 16.6 & 32.7 \\
    HunyuanVideo-1.5 (video)~\cite{wu2025hunyuanvideo} & 17.7 & 32.5 & 25.7 & 19.2 & 17.3 & 12.4 & 16.3 & 20.2 \\
    Qwen-Image-Edit-2511~\cite{wu2025qwen} & 15.7 & 16.9 & 34.2 & 23.2 & 17.1 & 13.4 & 17.2 & 19.7 \\
    BAGEL-7B-MoT~\cite{deng2025emerging} & 43.5 & 10.0 & 11.9 & 10.3 & 10.2 & 10.0 & 10.6 & 15.2 \\
    \bottomrule
  \end{tabular}
  \end{adjustbox}
\end{table}

The comparison reveals two main patterns. First, alignment is the bottleneck: models often produce plausible images that do not follow the visual page, and even GPT Image 2 drops to 20.0 on pose control. Second, V2V-Zero is strongest on attribute binding, reaching 76.9 on inline color, while structural categories remain weak (13.3 pose, 16.6 sketch). This suggests that the zero-shot ceiling is driven more by visual-structure conditioning than by semantic attribute binding.

The benchmark therefore exposes three tiers of visual-conditioning difficulty: attribute binding is already useful, content generation remains unreliable, and structural control remains the hardest regime. Figure~\ref{fig:v2vbench_qualitative}, placed early in the paper, shows the corresponding qualitative successes and failures.

\subsection{Validation on T2V}
\label{sec:hunyuan_video}

To test the T2V direction, we port the same visual-page conditioning path to HunyuanVideo-1.5~\cite{wu2025hunyuanvideo}. A Simple-V2V Bench page is encoded by the video model's Qwen2.5-VL multimodal encoder, padded or truncated to the pipeline's conditioning length, and used by the frozen temporal denoiser. We evaluate 616 videos (seven categories, four samples per prompt, 33 frames at 480p) with the same Qwen3-VL-32B judge applied directly to generated videos.

\begin{figure}[t]
  \centering
  \includegraphics[width=\linewidth,height=0.38\textheight,keepaspectratio]{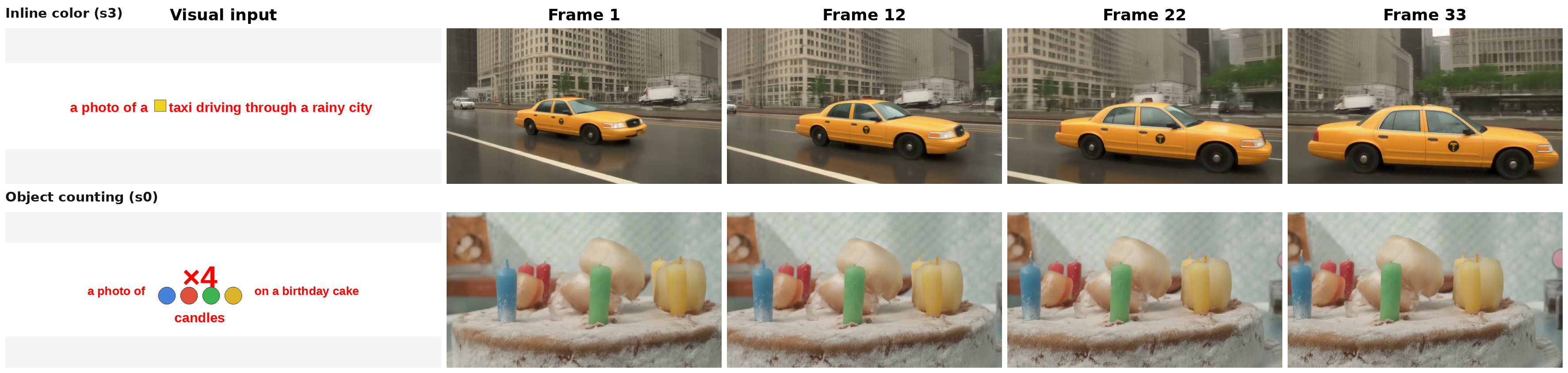}
  \caption{\textbf{HunyuanVideo-1.5 representative examples on Simple-V2V Bench.} Each row shows the visual input page and four uniformly sampled frames from one generated video. The examples illustrate inline-color and object-counting cases; the aggregate score of 20.2/100 in Table~\ref{tab:v2v_bench_baselines} is computed over all 616 generated videos using the VLM judge on the full video.}
  \label{fig:hunyuan_video_pilot}
\end{figure}

HunyuanVideo scores 20.2/100 overall (Table~\ref{tab:v2v_bench_baselines}), 12.5 points below V2V-Zero's image result. Inline color and inline visual reference transfer best, while pose and sketch remain near the floor. Figure~\ref{fig:hunyuan_video_pilot} shows representative inline-color and counting videos. This supports the architectural claim that a VLM-conditioned generator can consume visual pages without training, but a video-specific evaluation of temporal consistency and motion fidelity remains future work; per-category video details are in Appendix~\ref{app:hunyuan_video}.

\FloatBarrier

%==============================================================================
% SECTION 5: MECHANISM ANALYSIS
%==============================================================================

\section{Mechanism analysis}
\label{sec:mechanism}

This section isolates the evidence needed to explain why the V2V pathway works in practice. The central question is not whether the VLM can generate a useful textual rationale from a visual page, but whether the frozen generator actually uses the visual-page hidden states that V2V-Zero injects.

\subsection{The V2V architectural observation}
\label{sec:implicit_v2v_theorem}

Let a conventional T2I pipeline be written as $I=\mathcal{G}(\mathcal{E}(p))$, where $p$ is a user text prompt, $\mathcal{E}$ is the multimodal conditioning encoder, and $\mathcal{G}$ is a diffusion model that cross-attends directly to the encoder hidden states. If the same $\mathcal{E}$ also natively maps a visual page $V$ to hidden states in the same $D$-dimensional conditioning space, then replacing $p$ with $V$ yields the V2V-Zero route in Eq.~\ref{eq:pipeline}: the user input becomes visual, while the generator's learned conditioning interface and model weights are unchanged.

This observation is architectural rather than a formal theorem: it states when the V2V pathway exists. The experiments test whether the pathway is useful in practice, and show that exploiting it zero-shot yields substantial gains for visual specifications whose information is preserved in final-layer VLM hidden states.

\subsection{The reasoning path is visually routed}
\label{sec:hs_geometry}

On the full 553-prompt benchmark, the Qwen-Image-V2V-Zero \textsc{Full-Final} setting achieves 0.85 overall. We refer to this mode as the \emph{reasoning path}: the VLM first reads the visual page and generates scene-level reasoning tokens, and the final conditioning sequence concatenates visual-prefix hidden states with hidden states at the generated reasoning-token positions. Crucially, however, this name should not be read as a claim that V2V-Zero works by converting the visual page into a textual rationale. The reasoning path preserves the visual page hidden states themselves, and the generator can attend to them directly.

\begin{figure}[t]
\centering
\includegraphics[width=\linewidth]{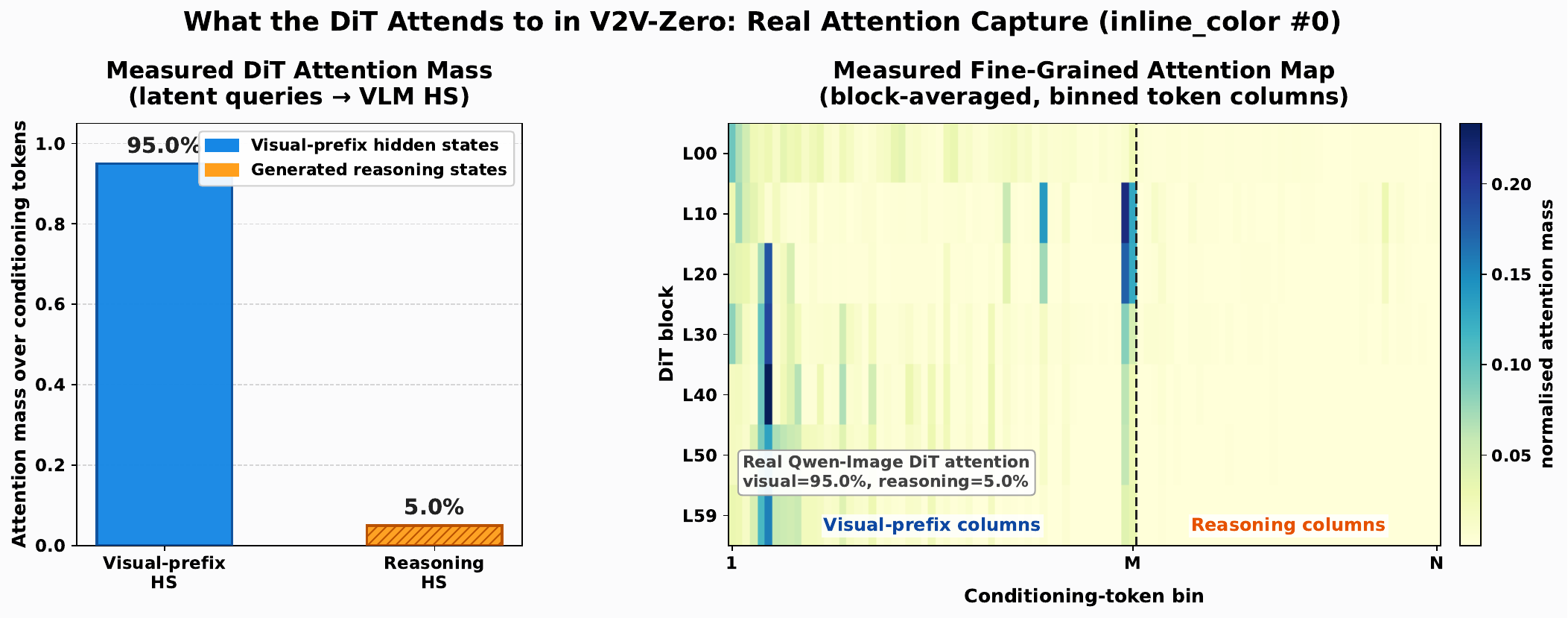}
\caption{\textbf{Real DiT attention routing in the V2V-Zero reasoning path.}
We hook Qwen-Image DiT joint attention during a real inline-color V2V-Bench generation and measure attention from latent image queries to VLM conditioning hidden states. The \textsc{Full-Final} reasoning path contains both visual-prefix states from the visual page and generated reasoning-token states, but the DiT assigns 95.0\% of conditioning-token attention to the visual-prefix states and only 5.0\% to the generated reasoning states. Since the visual prefix accounts for 266 of 466 conditioning tokens (57.1\%), the measured distribution shows that the generator preferentially routes through visual-page hidden states rather than relying on generated reasoning tokens.}
\label{fig:dit_attention_real}
\end{figure}

To test the actual routing used by the generator, we instrument Qwen-Image DiT attention during a real V2V-Bench run. Specifically, for selected DiT blocks, we measure the softmax attention mass from latent image queries to VLM conditioning-token key columns, and split those conditioning columns into visual-prefix hidden states and generated reasoning-token hidden states. Figure~\ref{fig:dit_attention_real} shows the result for the inline-color sample used in the qualitative comparison. The conditioning sequence contains 266 visual-prefix states and 200 generated reasoning states. If attention were merely proportional to token count, the visual prefix would receive 57.1\% of conditioning-token mass. Instead, the DiT assigns 95.0\% of the measured conditioning-token attention to the visual-prefix hidden states and only 5.0\% to the generated reasoning states. This is a direct routing measurement from the actual zero-shot V2V path: the generator overwhelmingly reads from the visual page representation rather than from the generated reasoning-token representation.

This finding is central to the mechanism. V2V-Zero does not succeed because the VLM produces a useful textual rationale that replaces the visual input. It succeeds because the pretrained DiT can consume final-layer VLM hidden states whose visual-prefix portion remains effective as generation conditioning, even though the model was not fine-tuned for visual-page prompting. The generated reasoning tokens make the sequence compatible with the generator's language-conditioned interface, but the dominant information pathway is hidden-state visual conditioning. Fixed interpreter instructions help place the page into the expected conditioning format, but the measured routing shows that the DiT primarily reads visual-page hidden states rather than generated reasoning-token states. This supports the paper's main mechanistic claim: V2V-Zero exposes a latent visual-to-visual route already present in VLM-conditioned generators, beyond text re-encoding.

Only the final VLM layer is generation-compatible: injecting from layer $L{-}1$ drops to 5.39\% despite achieving the highest text-visual alignment ($\rho{=}0.908$ vs.\ $\rho_L{=}0.712$), because only layer $L$ matches the DiT's trained conditioning distribution. The generated reasoning tokens complement but do not replace visual-prefix states as the main conditioning source---Image-HS-only (71.57\%) already outperforms reasoning-only injection (68.14\%), confirming that visual states carry the dominant signal. Extended layer-wise analysis, token-level alignment diagnostics, and information-theoretic bottleneck analysis are provided in Appendix~\ref{app:hs_geometry}.

% \clearpage

%==============================================================================
% SECTION 6: CONCLUSION
%==============================================================================

\section{Conclusion}
\label{sec:conclusion}

We have proposed visual-to-visual (V2V) generation as a paradigm in which structured visual inputs serve as the primary conditioning signal for generative models. V2V-Zero is a zero-shot instantiation: by routing visual-page hidden states through the frozen VLM encoder of an existing T2I system, it reaches 0.85 on GenEval without fine-tuning, demonstrating that user-provided visual evidence can supply conditioning information that text-only prompts compress away. Simple-V2V Bench reveals a three-tier capability structure---attribute binding approaches commercial baselines, content generation shows a clear gap, and structural control remains open---while the HunyuanVideo-1.5 extension confirms that the same interface transfers to video generation.

More broadly, V2V reframes visual input as a first-class conditioning language for generative models. Text scaling has driven recent progress, but visual specification offers a complementary axis: layout, color, identity, and structure are often specified more directly as images than as serialized descriptions. Native V2V models trained end-to-end, richer specification formats, and evaluation frameworks that reward precise visual instruction following are the natural next steps along this axis.

%==============================================================================
% REFERENCES
%==============================================================================

\clearpage
\bibliographystyle{unsrtnat}
\bibliography{references}

%%%%%%%%%%%%%%%%%%%%%%%%%%%%%%%%%%%%%%%%%%%%%%%%%%%%%%%%%%%%

\clearpage
\appendix

% =============================================================================
% Appendix
% =============================================================================

% Let appendix floats share pages more freely; otherwise short appendix sections
% can flush figures/tables onto sparse float-only pages.
\setcounter{topnumber}{5}
\setcounter{bottomnumber}{5}
\setcounter{totalnumber}{8}
\renewcommand{\topfraction}{0.95}
\renewcommand{\bottomfraction}{0.85}
\renewcommand{\textfraction}{0.05}
\renewcommand{\floatpagefraction}{0.80}

\section{Mechanism and Signal Diagnostics}
\label{app:hs_geometry}
\label{app:bottleneck}
\label{app:information_advantage}

The main mechanism claim is empirical: the default \textsc{Full-Final}
reasoning path is primarily visually routed. In the real attention capture in
Section~\ref{sec:mechanism}, the Qwen-Image DiT assigns 95.0\% of measured
conditioning-token attention mass to visual-prefix hidden states and 5.0\% to
generated reasoning states. The diagnostics below explain why this routing is
plausible: image-token states are not merely text paraphrases, while only the
final VLM layer is distributionally compatible with the DiT conditioning
interface.

\begin{table}[htbp]
  \centering
  \caption{Hidden-state diagnostics.}
  \label{tab:hidden_state_diagnostics_appendix}
  \small
  \begin{adjustbox}{width=\textwidth}
  \begin{tabular}{lll}
    \toprule
    \textbf{Diagnostic} & \textbf{Value} & \textbf{Interpretation} \\
    \midrule
    Image-token vs text hidden-state cosine & 0.64 & Visual states differ from text paraphrases \\
    Reasoning-token vs text hidden-state cosine & 0.96 & VLM reasoning is close to text conditioning \\
    Layer $L{-}1$ visual-text alignment & 0.908 & Strong semantic alignment but not generation-compatible \\
    Final layer $L$ visual-text alignment & 0.712 & Lower alignment but matches DiT training interface \\
    GenEval layer $L$ injection & 86.77\% & Validated on diagnostic subset \\
    GenEval layer $L{-}1$ injection & 5.39\% & Off-distribution for the DiT \\
    \bottomrule
  \end{tabular}
  \end{adjustbox}
\end{table}

\begin{table}[htbp]
  \centering
  \caption{Demoted retrieval controls for the rendered text-page diagnostic. Token-max is the main proof in Section~\ref{sec:mechanism}; these controls show that the effect weakens under pooling and does not transfer to pure color pages.}
  \label{tab:textpage_retrieval_controls}
  \small
  \begin{adjustbox}{width=\textwidth}
  \begin{tabular}{lcccc}
    \toprule
    \textbf{Diagnostic} & \textbf{R@1} & \textbf{R@3} & \textbf{MRR} & \textbf{Margin} \\
    \midrule
    Text-page token-max (main proof) & 0.68 & 0.84 & 0.773 & 0.078 \\
    Text-page mean-pool control & 0.20 & 0.44 & 0.409 & 0.047 \\
    Text-page full-token control & 0.08 & 0.44 & 0.320 & 0.017 \\
    Color-page token-max control & 0.0625 & 0.0625 & 0.182 & -0.0047 \\
    Color-page mean-pool control & 0.0625 & 0.25 & 0.237 & 0.0012 \\
    \bottomrule
  \end{tabular}
  \end{adjustbox}
\end{table}

\subsection{Layer sensitivity}
\label{sec:layerwise}

Direct injection is extremely sensitive to layer choice. Using the final VLM layer reaches 86.77\% on the GenEval ablation subset, while injecting from layer $L{-}1$ drops to 5.39\%. This is expected: during T2I training, the DiT learns to consume the encoder's final-layer conditioning distribution, so only layer $L$ representations reliably lie in its learned interface. Earlier layers, while semantically rich, lie outside this distribution. Notably, $L{-}1$ achieves \emph{the highest} text-visual alignment ($\rho{=}0.908$), surpassing the final layer ($\rho_{L}{=}0.712$)---a design insight for future systems that could train the DiT to accept earlier-layer states.

\subsection{Reasoning tokens and pre-alignment}
\label{sec:mismatch}
\label{sec:bottleneck}
\label{sec:information_advantage}
\label{sec:pre_alignment}

Image-HS-only and \textsc{Full-Final} isolate two parts of the same direct-conditioning path. Image-HS-only preserves visual evidence from the page, including count, layout, and color. \textsc{Full-Final} keeps that evidence and adds final-layer reasoning states that align the page with a generation-ready scene interpretation. The attention evidence in Figure~\ref{fig:dit_attention_real} clarifies the role of those reasoning states: they do not replace the visual page as the main conditioning source. Instead, the reasoning path remains visually routed, with the DiT assigning most conditioning-token attention to the visual-prefix hidden states.

\begin{figure}[!tbp]
\centering
\includegraphics[width=0.68\linewidth]{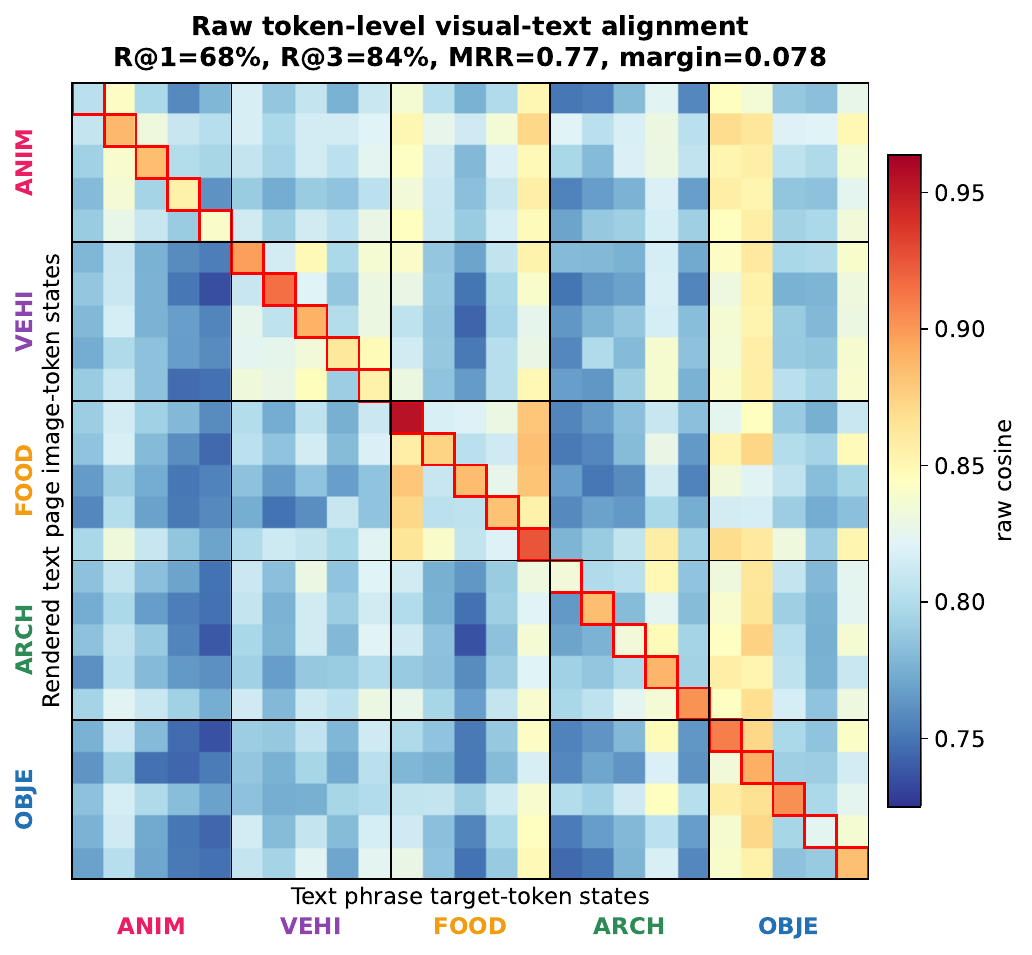}
\caption{\textbf{Token-level cross-modal alignment on rendered text pages.}
Rendered text-page image-token states retrieve their matching phrase-token states with $R@1{=}68\%$, $R@3{=}84\%$, and $\mathrm{MRR}{=}0.773$, showing local visual-text alignment in the injected VLM hidden states.}
\label{fig:tokenmax_alignment}
\end{figure}

The token-max diagnostic in Figure~\ref{fig:tokenmax_alignment} adds the complementary observation that those visual-prefix states are locally aligned with the corresponding phrase tokens. Together these results explain why the default mode improves over the non-reasoning control while still preserving the direct visual-state interface.

\section{Detailed Related Work}
\label{app:related_work}

\noindent \textbf{Text-first visual generation and specialized editing.}
T2I and T2V models have advanced rapidly by scaling diffusion, transformer, and multimodal-conditioning backbones~\cite{ddpm,ldm,imagen,sdxl,peebles2023scalable,liu2024redefining,wu2025qwen,makeavideo,yang2024cogvideox,liupusa,polyak2024movie,chen2023videocrafter1,wu2025hunyuanvideo,liu2024evalcrafter}. Most of these systems still expose natural language as the default user interface, even when the generator internally relies on strong encoders such as T5, CLIP, or VLMs~\cite{raffel2020exploring,clip,bai2025qwen25vltechnicalreport}. A parallel line studies image and video editing, where the input visual content is treated as a source to modify and the user intent is usually specified by text, inversion, attention control, instruction tuning, or temporal feature propagation~\cite{meng2021sdedit,hertz2022prompt,brooks2023instructpix2pix,geyer2023tokenflow,ku2024anyv2v}. These works provide strong text-guided synthesis and editing capabilities, but they largely preserve separate operational modes: T2I, T2V, image editing, video editing, and image-to-video are optimized and evaluated as different tasks.

\noindent \textbf{Visual conditions as task-specific mechanisms.}
The recurring limitation of text-only prompting is that spatial layout, appearance, identity, typography, color, and motion are intrinsically visual. Prior work therefore adds visual conditions through specialized channels: ControlNet and T2I-Adapter inject structural maps, GLIGEN grounds generation with boxes and other spatial signals, IP-Adapter and DreamBooth use reference images for appearance or identity, Stable Video Diffusion and VideoComposer condition video generation on images, sketches, motion, or reference videos, and recent all-in-one systems such as OmniGen, OmniGen2, and VACE combine multiple generation and editing tasks in one framework~\cite{zhang2023adding,mou2024t2i,li2023gligen,ye2023ip,ruiz2023dreambooth,blattmann2023stable,wang2023videocomposer,xiao2025omnigen,wu2026omnigen2instructionalignedmultimodalgeneration,jiang2025vace}. Text rendering makes the boundary between language and vision especially explicit: character-aware models like GlyphControl, AnyText, and TextDiffuser show that words can be treated as glyphs, layouts, and pixels rather than only as discrete prompt tokens~\cite{liu2023character,yang2023glyphcontrol,tuo2023anytext,chen2023textdiffuser,betker2023improving,wu2025qwen}. These methods demonstrate the value of visual evidence, but the evidence is usually tied to a predefined adapter, control map, edit input, or benchmark category.

\noindent \textbf{Visual context and native multimodal generation.}
The closest conceptual line treats images themselves as prompts or in-context demonstrations. Visual Prompting via Image Inpainting, Prompt Diffusion, Context Diffusion, Stable Diffusion visual in-context learning, VisualCloze, RealGeneral, and UNIC all study visual examples, context, or token sequences as a route to more general image and video generation/editing~\cite{bar2022visual,wang2023context,najdenkoska2024context,oorloff2025stable,li2025visualcloze,lin2025realgeneral,ye2025unic}. In parallel, native or unified multimodal generators such as Emu3.5, Show-o2, BAGEL, Tuna-2, and DeepGen~1.0 collapse understanding and generation into shared or tightly coupled architectures~\cite{cui2025emu3,xie2025show,deng2025emerging,liu2026tuna,wang2026deepgen}. These works are important evidence that visual generation is moving beyond pure language prompting. However, most still inherit conventional task definitions and benchmarks: T2I, image editing, text rendering, subject/reference generation, video generation, or video editing. V2V instead studies \emph{visual context itself} as the user interface: text can be rendered into vision, and color swatches, glyphs, sketches, object thumbnails, style references, layouts, and temporal cues can coexist on the same specification page. V2V-Zero asks whether existing VLM-conditioned generators already expose this interface without training, and Simple-V2V Bench measures the resulting capability gap directly.

\section{Reproducibility and Implementation Details}
\label{app:implementation}
\label{app:implementation_rendering}

This appendix is intended as an evidential supplement to the main paper. It
therefore emphasizes reproducibility, exact evaluation scope, and additional
quantitative and qualitative evidence.

\subsection{V2V-Zero inference}
\label{app:algorithm}

\begin{algorithm}[htbp]
\caption{V2V-Zero inference in the validated \textsc{Full-Final} path}
\label{alg:v2vzero_appendix}
\begin{algorithmic}[1]
\REQUIRE Visual specification page $V$, fixed template $T$, frozen VLM encoder $\mathcal{E}$, frozen generator $\mathcal{G}$, reasoning-token budget $N$
\ENSURE Generated image $I$
\STATE Compose the multimodal VLM input $\mathbf{x}=[t_{\mathrm{sys}}, \mathrm{ViT}(V), t_{\mathrm{user}}(T), \texttt{<gen>}]$
\STATE Generate reasoning tokens $[t_1,\ldots,t_N] \gets \mathcal{E}_{\mathrm{AR}}(\mathbf{x})$ with greedy decoding
\STATE Recompute final-layer states $\tilde{H}\gets\mathcal{E}^{(L)}([\mathbf{x},t_1,\ldots,t_N])$
\STATE Extract $H_{\mathrm{img}}=\tilde{H}_{\operatorname{ViT}(V)}$ and $H_{\mathrm{reason}}=\tilde{H}_{-N:}$
\STATE Form the generator conditioning sequence $H=[H_{\mathrm{img}};H_{\mathrm{reason}}]$
\STATE Replace the generator prompt embedding with $H$
\STATE Sample $I\gets\mathcal{G}(H)$ with the frozen denoiser
\RETURN $I$
\end{algorithmic}
\end{algorithm}

The visual page $V$ is the user-controlled input and is never used as an edit
image. It is read only by the VLM vision encoder $\mathcal{E}$, and the
generator receives hidden states rather than VAE latents of the page. This
distinction is important: the page is a specification document, not a source
image to reconstruct. The template $T$ is fixed implementation guidance for
interpreting $V$, not an additional user prompt in the V2V formulation.
V2V-Zero therefore requires implementation access to internal conditioning
states; it is not equivalent to calling an unmodified black-box T2I API with an
image replacing a string. It changes only inference-time routing: exposing VLM
hidden states and replacing prompt embeddings.

\subsection{Conditioning modes}
\label{app:modes}

\begin{table}[htbp]
  \centering
  \caption{Conditioning modes used in the paper. \textsc{Full-Final} is the
  default reasoning path; Image-HS only is the non-reasoning visual-state
  control.}
  \label{tab:appendix_modes}
  \small
  \begin{adjustbox}{width=\textwidth}
  \begin{tabular}{llll}
    \toprule
    \textbf{Mode} & \textbf{Image states} & \textbf{Reasoning states} & \textbf{Role} \\
    \midrule
    \textsc{Full-Final} & \checkmark & \checkmark & Default reasoning path \\
    Image-HS only & \checkmark & -- & Non-reasoning control \\
    \bottomrule
  \end{tabular}
  \end{adjustbox}
\end{table}

Relative to Image-HS only, \textsc{Full-Final} has higher inference overhead
because it autoregressively generates reasoning tokens, performs one
teacher-forced VLM recomputation pass, and conditions the generator on a longer
sequence. This extra compute corresponds to the diagnostic-subset improvement
from 71.57\% to 86.77\%.

\subsection{Visual page families}
\label{app:page_taxonomy}

\begin{table}[htbp]
  \centering
  \caption{Visual page families used across GenEval, text rendering, and
  Simple-V2V Bench. The important design rule is that the page should encode the
  intended scene rather than invite the VLM to describe the document itself.}
  \label{tab:page_taxonomy}
  \small
  \begin{tabular}{lll}
    \toprule
    \textbf{Family} & \textbf{Visual evidence} & \textbf{Primary use} \\
    \midrule
    Color card & Solid RGB swatches & Color binding \\
    Object layout card & Thumbnails in spatial cells & Object relation and position \\
    Counting display & Repeated thumbnails or marks & Exact object count \\
    Inline color prompt & 28px swatch inside prompt text & Token-level color binding \\
    Inline visual reference & 72px object thumbnail inside prompt text & Subject identity \\
    Rendered text page & Target glyphs and font style & Text rendering \\
    Style reference page & Artwork or visual style image & Style transfer \\
    Structure reference page & Sketch or pose skeleton & Spatial/pose control \\
    \bottomrule
  \end{tabular}
\end{table}

\begin{figure}[!tbp]
  \centering
  \includegraphics[width=\linewidth]{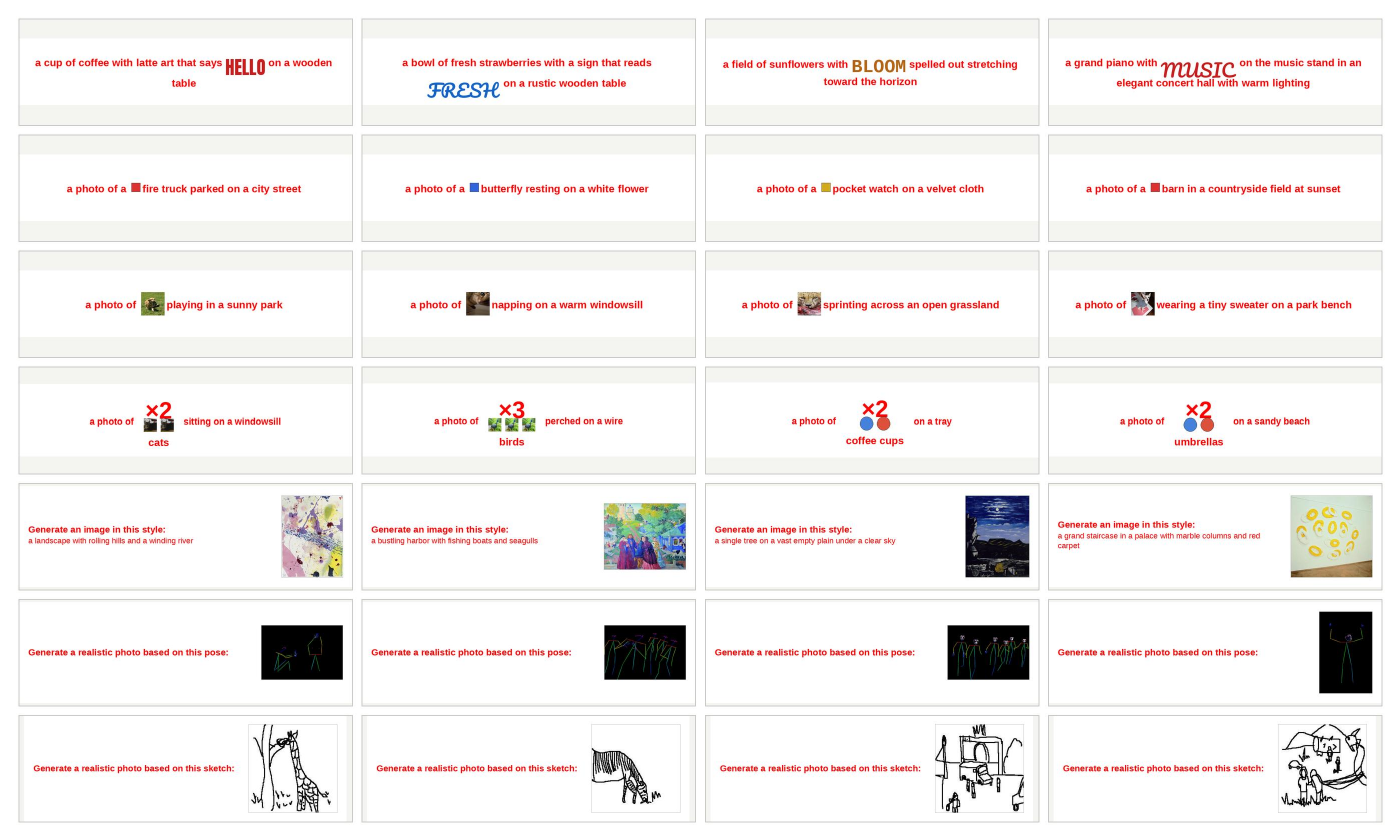}
  \caption{\textbf{Simple-V2V Bench visual-page atlas.} Representative input
  pages from the seven task families show the visual evidence that models must
  read from the page, including rendered text, inline swatches, visual
  references, counting displays, style references, pose skeletons, and sketches.}
  \label{fig:appendix_v2vbench_page_atlas}
\end{figure}

\noindent \textbf{Core settings.}
GenEval V2V-Zero image experiments use Qwen2.5-VL-7B-Instruct as the VLM
conditioning backbone and Qwen-Image-2512 as the frozen image generator. Unless
otherwise specified, decoding is greedy and each GenEval prompt is sampled four
times. Simple-V2V Bench generation uses 1024$\times$1024 outputs, 30 denoising
steps, CFG scale 4.0, seed 42, 200 reasoning tokens, and \textsc{Full-Final}
mode for all seven categories.

\section{GenEval Evaluation Details}
\label{app:geneval_details}

\subsection{Evaluation protocol}

\begin{table}[htbp]
  \centering
  \caption{GenEval result scopes. Fullbench results use all 553 prompts and
  2212 generated images; subset results use 102 prompts and 408 images and are
  diagnostic only.}
  \label{tab:geneval_protocol}
  \small
  \begin{adjustbox}{width=\textwidth}
  \begin{tabular}{lllll}
    \toprule
    \textbf{Scope} & \textbf{Prompts} & \textbf{Samples/prompt} & \textbf{Images} & \textbf{Use in paper} \\
    \midrule
    Fullbench & 553 & 4 & 2212 & Main comparison \\
    Ablation subset & 102 & 4 & 408 & Two-mode and hyperparameter diagnostics \\
    Incomplete fullbench & $<553$ & varies & varies & Excluded from claims \\
    Broken-checkpoint runs & varies & varies & varies & Excluded \\
    \bottomrule
  \end{tabular}
  \end{adjustbox}
\end{table}

\section{GenEval Ablations and Diagnostics}
\label{app:full_ablation}
\label{app:token_sweep}
\label{app:ablation_other}

All ablations in this section use the GenEval diagnostic subset (102 prompts, 4 samples each) unless noted otherwise. The default configuration is \textsc{Full-Final} mode with 150 reasoning tokens, font size 32px, and text-to-image height ratio 0.20. On the full 553-prompt benchmark, 200 tokens performs slightly better (0.85 vs.\ 0.84 for 150 tokens), so the main-body result uses 200 tokens; the ablation subset favors 150 tokens.

\subsection{Token count sweep}
\label{tab:full_ablation_anchor}

Table~\ref{tab:token_batch} sweeps the reasoning-token budget under the default \textsc{Full-Final} mode, and includes two ablation baselines: Image-HS-only (visual hidden states without reasoning) and Reason-only (reasoning-token states without image hidden states). Performance peaks at 150 tokens on the diagnostic subset; 200 tokens is marginally better on the full benchmark (0.85 vs.\ 0.84). The baselines isolate the contribution of each conditioning pathway: Image-HS-only scores 71.57\% and Reason-only scores 68.14\%, while \textsc{Full-Final} at 150 tokens reaches 86.77\%, confirming that both pathways contribute and their combination is synergistic (+15.20\,pp over Image-HS-only, +18.63\,pp over Reason-only).

\begin{table}[htbp]
\centering
\caption{\textbf{Token count sweep and conditioning-mode ablation} (diagnostic subset, 102 prompts $\times$ 4 samples). \textsc{Full-Final} (FF) rows sweep the reasoning-token budget; Image-HS-only uses visual hidden states without reasoning; Reason-only uses 200 generated reasoning-token states without image hidden states. Best per-column in \textbf{bold}.}
\label{tab:token_batch}
\label{tab:full_ablation}
\small
\begin{tabular}{@{}rlccccccc@{}}
\toprule
Mode & Tokens & Overall & Single & TwoObj & Counting & Colors & Position & ColAttr \\
\midrule
FF & 50  & 82.11 & 98.53 & 94.12 & 80.88 & 92.65 & 60.29 & 66.18 \\
FF & 100 & 84.56 & 98.53 & 91.18 & 89.71 & \textbf{94.12} & 66.18 & 67.65 \\
\textbf{FF} & \textbf{150} & \textbf{86.77} & \textbf{98.53} & \textbf{94.12} & \textbf{89.71} & 88.24 & \textbf{75.00} & \textbf{75.00} \\
FF & 200 & 84.31 & 98.53 & 92.65 & 86.76 & 88.24 & 72.06 & 67.65 \\
FF & 300 & 84.80 & 98.53 & 91.18 & 82.35 & 88.24 & 75.00 & 73.53 \\
\midrule
Image-HS & --- & 71.57 & 98.53 & 76.47 & 80.88 & 89.71 & 35.29 & 48.53 \\
Reason-only & 200 & 68.14 & 67.65 & 80.88 & 73.53 & 83.82 & 45.59 & 57.35 \\
\bottomrule
\end{tabular}
\end{table}

\subsection{Rendering parameter ablations}

Table~\ref{tab:rendering_full} ablates the visual-page rendering parameters: font size and text-to-image height ratio. Font 32px and ratio 0.20 are the validated defaults; deviations in either direction degrade performance, with large fonts (56--72px) causing catastrophic drops.

\begin{table}[htbp]
\centering
\caption{\textbf{Rendering parameter ablations} (\textsc{Full-Final}, 150 tokens). Default values in \textbf{bold rows}. $\Delta$ is relative to default.}
\label{tab:rendering_full}
\small
\begin{tabular}{@{}lcccccr@{}}
\toprule
Parameter & Overall & Position & ColAttr & Counting & Colors & $\Delta$ Overall \\
\midrule
\multicolumn{7}{l}{\emph{Font size (px)}} \\
Font 24 & 84.80 & 52.94 & \textbf{82.35} & 89.71 & 88.24 & $-$1.97 \\
\textbf{Font 32 (default)} & \textbf{86.77} & \textbf{75.00} & 75.00 & 89.71 & 88.24 & --- \\
Font 44 & 69.12 & 26.47 & 44.12 & 80.88 & 89.71 & $-$17.65 \\
Font 48 & 80.88 & 61.76 & 69.12 & 75.00 & 92.65 & $-$5.89 \\
Font 56 & 60.05 & 23.53 & 16.18 & 67.65 & 82.35 & $-$26.72 \\
Font 72 & 55.15 & 30.88 & 10.29 & 72.06 & 69.12 & $-$31.62 \\
\midrule
\multicolumn{7}{l}{\emph{Text-to-image height ratio}} \\
Ratio 0.15 & 84.31 & 57.35 & 77.94 & 88.24 & 88.24 & $-$2.46 \\
\textbf{Ratio 0.20 (default)} & \textbf{86.77} & \textbf{75.00} & 75.00 & 89.71 & 88.24 & --- \\
Ratio 0.30 & 76.96 & 45.59 & 64.71 & 82.35 & 85.29 & $-$9.81 \\
\bottomrule
\end{tabular}
\end{table}

\noindent\textbf{Key observations.}
(1)~Token count is non-monotonic: 150 tokens is the diagnostic-subset optimum, but the curve is relatively flat between 100--300 tokens (82--87\%). The two ablation baselines reveal complementary roles: Image-HS-only (71.57\%) retains strong Single-object and Colors performance but collapses on Position (35.29\%) and ColAttr (48.53\%); Reason-only (68.14\%) partially recovers Position (45.59\%) and ColAttr (57.35\%) but loses Single-object recognition (67.65\%). \textsc{Full-Final} combines both pathways synergistically (+15.20\,pp over Image-HS, +18.63\,pp over Reason-only).
(2)~Font 32px is a sharp optimum; font 24 trades position accuracy for color attribution, while fonts $\geq$44px cause catastrophic position and attribute drops.
(3)~Text ratio 0.20 is a sharp optimum; 0.30 loses nearly 10pp overall.

\section{Simple-V2V Bench Construction}
\label{app:simple_v2v_construction}

Simple-V2V Bench evaluates whether a model can read a visual specification page
and generate the corresponding image. It contains seven categories, 22 prompts
per category, and four generated samples per prompt, for 616
outputs per complete model run.

\begin{table}[htbp]
  \centering
  \caption{Simple-V2V Bench category construction.}
  \label{tab:simple_v2v_construction}
  \small
  \begin{tabular}{lll}
    \toprule
    \textbf{Category} & \textbf{Page format} & \textbf{Evaluated capability} \\
    \midrule
    Visual text & Styled target word inside scene prompt & Text rendering and style \\
    Inline color & Color swatch replaces color word & Color binding \\
    Inline visual reference & Object thumbnail replaces subject noun & Visual subject identity \\
    Object counting & Repeated objects or marks plus text & Count following \\
    Style transfer & Text instruction plus artwork reference & Style matching \\
    Pose control & OpenPose skeleton plus minimal text & Human pose structure \\
    Sketch reference & Line drawing plus minimal text & Layout and structure \\
    \bottomrule
  \end{tabular}
\end{table}

The final V2V-Zero run uses Qwen2.5-VL-7B-Instruct plus Qwen-Image-2512,
\textsc{Full-Final} mode,
30 denoising steps, CFG scale 4.0, seed 42, and 200 reasoning tokens.

\section{Simple-V2V Bench Evaluation Protocol}
\label{app:simple_v2v_protocol}

The evaluator is a direct two-image judge. Qwen3-VL-32B-Instruct
receives the input page and one generated output image, then independently
scores two dimensions: \emph{Quality} (1--10) and \emph{Alignment} (1--10).
The final score is $\min(\text{Quality}, \text{Alignment}) \times 10$, a
bottleneck formula that ensures one weak dimension drags down the overall
result. Scores are averaged over the four samples for each prompt, then
averaged over prompts and categories. This is a mean-of-samples protocol, not
an oracle or best-of-four protocol.

\begin{table}[htbp]
  \centering
  \caption{Simple-V2V Bench scoring dimensions.}
  \label{tab:simple_v2v_rubric}
  \small
  \begin{tabular}{lll}
    \toprule
    \textbf{Dimension} & \textbf{Range} & \textbf{Meaning} \\
    \midrule
    Quality & 1--10 & Image fidelity, coherence, absence of artifacts \\
    Alignment & 1--10 & Faithfulness to the visual specification on the input page \\
    \midrule
    Final score & 10--100 & $\min(\text{Quality}, \text{Alignment}) \times 10$ \\
    \bottomrule
  \end{tabular}
\end{table}

The judge follows strict per-category rubrics with explicit deduction amounts
(e.g., wrong object count $= -3$ alignment, color mismatch $= -4$ alignment).
Before assigning scores, the judge must enumerate all flaws in its analysis.
Calibration anchors define each score level (e.g., alignment 8--9 means ``minor
imperfections only''; alignment 1--2 means ``core specification missing or
unusable''). The judge tone is ``ruthlessly strict''---it never gives the
benefit of the doubt.

\section{Simple-V2V Bench Extended Results}
\label{app:simple_v2v_results}

\begin{table}[htbp]
  \centering
  \caption{V2V-Zero results on Simple-V2V Bench. Quality (Q) and Alignment (A) are each scored 1--10; final score $= \min(\text{Q},\text{A})\times 10$.}
  \label{tab:simple_v2v_v2vzero_appendix}
  \small
  \begin{tabular}{lcccc}
    \toprule
    \textbf{Category} & \textbf{Score /100} & \textbf{Quality} & \textbf{Alignment} \\
    \midrule
    Inline color & 76.93 & 9.51 & 7.69 \\
    Inline visual reference & 42.84 & 8.01 & 4.28 \\
    Visual text & 34.77 & 6.58 & 3.48 \\
    Object counting & 23.98 & 5.39 & 2.40 \\
    Style transfer & 20.34 & 7.58 & 2.03 \\
    Sketch reference & 16.59 & 5.44 & 1.66 \\
    Pose control & 13.30 & 2.98 & 1.33 \\
    \midrule
    \textbf{Overall} & \textbf{32.68} & 6.50 & 3.27 \\
    \bottomrule
  \end{tabular}
\end{table}

\begin{table}[htbp]
  \centering
  \caption{Simple-V2V Bench comparison (same dual-dimension protocol). All scores use
  dual-dimension scoring ($\min(\text{Q},\text{A})\times 10$). HunyuanVideo is
  evaluated by applying the VLM judge to the generated video directly.}
  \label{tab:simple_v2v_full_comparison_appendix}
  \small
  \setlength{\tabcolsep}{3pt}
  \begin{adjustbox}{width=\textwidth}
  \begin{tabular}{lcccccccc}
    \toprule
    \textbf{Model} & \textbf{Vis.Text} & \textbf{Inl.Color} & \textbf{Inl.Ref} & \textbf{Count} & \textbf{Style} & \textbf{Pose} & \textbf{Sketch} & \textbf{Overall} \\
    \midrule
    GPT Image 2 & 78.30 & \textbf{92.39} & 75.80 & \textbf{91.82} & \textbf{60.34} & \textbf{20.00} & \textbf{33.98} & \textbf{64.66} \\
    Seedream 5.0 Lite & \textbf{78.98} & 68.67 & 74.66 & 88.79 & 48.71 & 16.78 & 32.42 & 58.43 \\
    Nano Banana 2 & 59.20 & 69.66 & \textbf{77.95} & 67.05 & 44.66 & 19.09 & 22.27 & 51.41 \\
    V2V-Zero & 34.77 & 76.93 & 42.84 & 23.98 & 20.34 & 13.30 & 16.59 & 32.68 \\
    HunyuanVideo-1.5 & 17.73 & 32.50 & 25.68 & 19.20 & 17.27 & 12.39 & 16.25 & 20.15 \\
    Qwen-Image-Edit-2511 & 15.68 & 16.93 & 34.20 & 23.18 & 17.05 & 13.41 & 17.16 & 19.66 \\
    BAGEL-7B-MoT & 43.52 & 10.00 & 11.93 & 10.34 & 10.23 & 10.00 & 10.57 & 15.23 \\
    \bottomrule
  \end{tabular}
  \end{adjustbox}
\end{table}

\begin{figure}[!tbp]
  \centering
  \includegraphics[width=\linewidth]{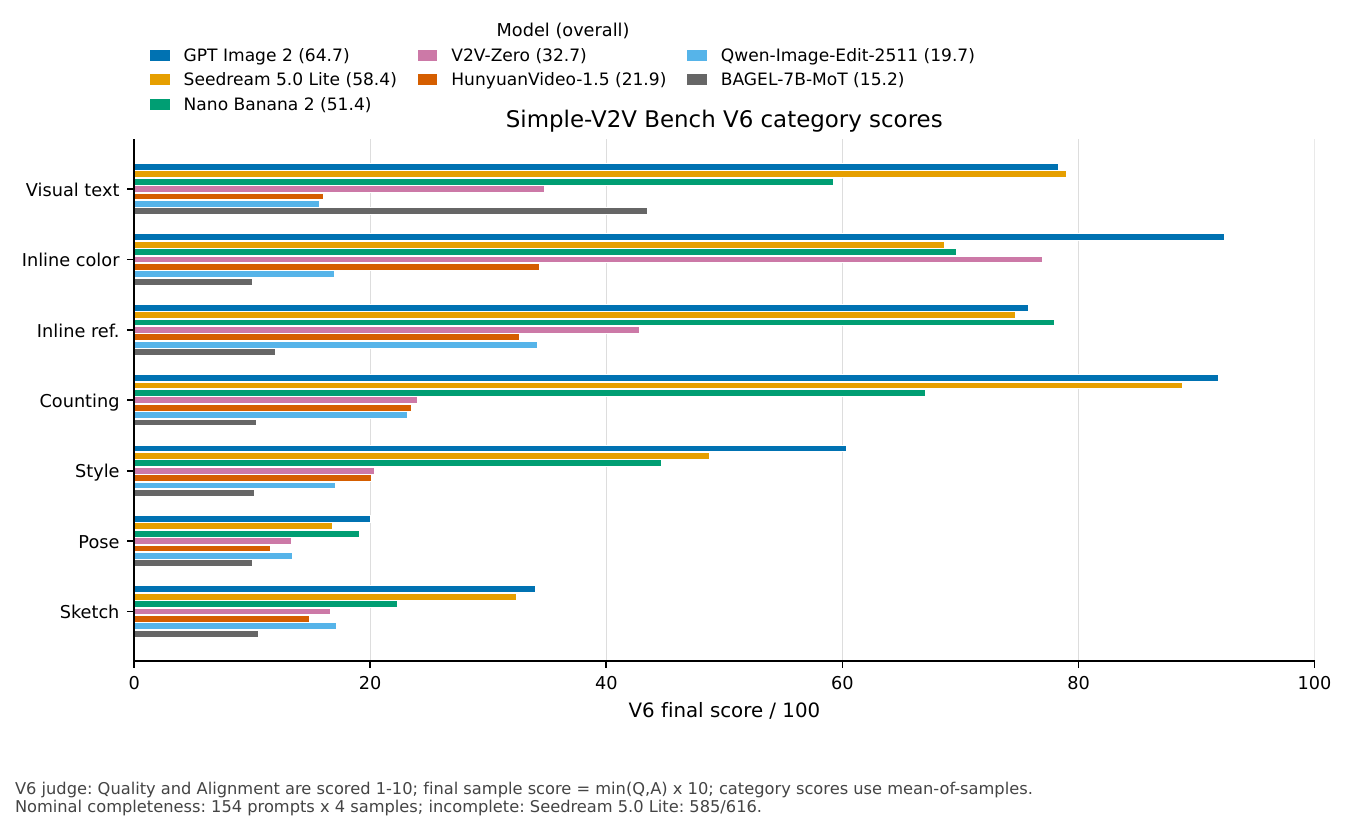}
  \caption{\textbf{Simple-V2V Bench category scores.} Category-level scores
  expose which visual specification types are handled reliably and which remain
  difficult across models. The strongest systems remain much weaker on pose and
  sketch control than on inline color, visual reference, and object counting.}
  \label{fig:appendix_v2vbench_v6_category_bars}
\end{figure}

\begin{figure}[!tbp]
  \centering
  \includegraphics[width=\linewidth]{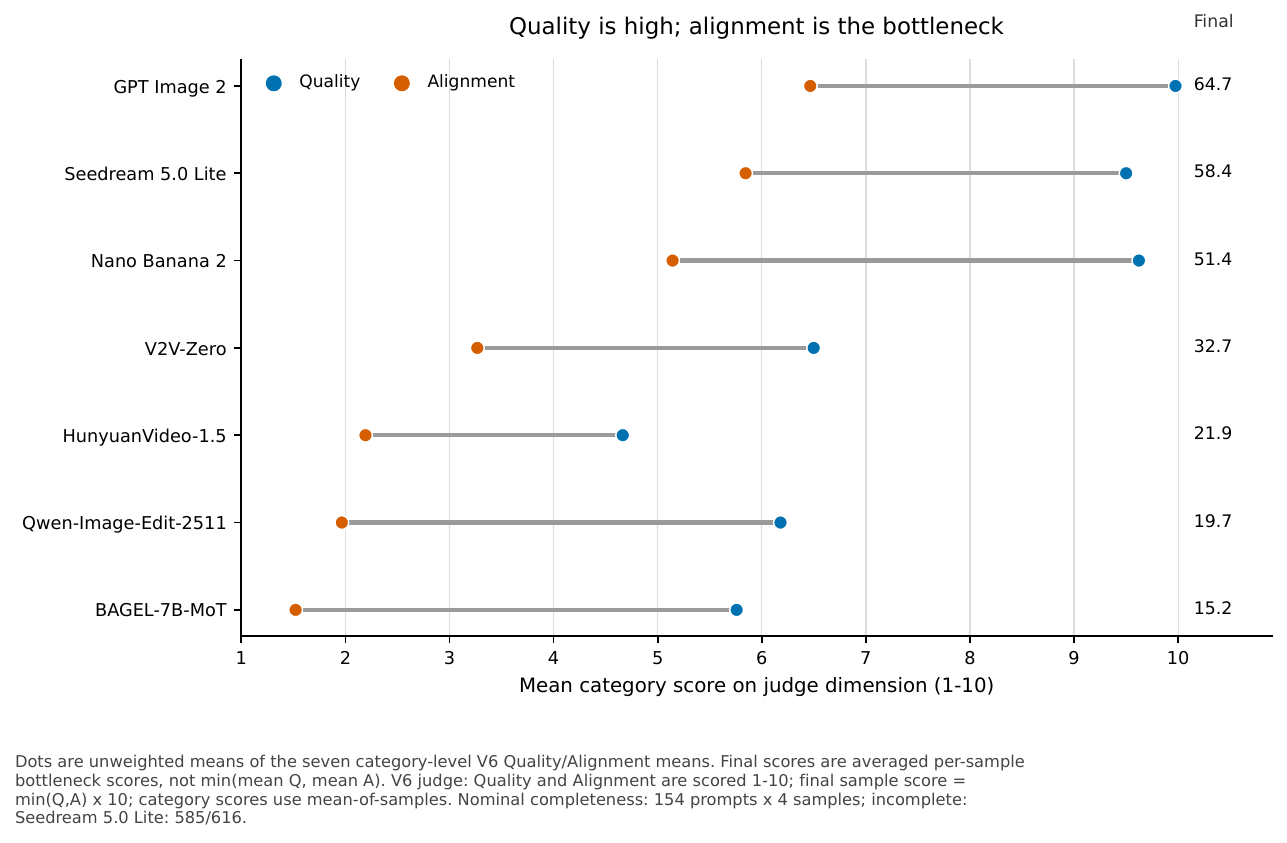}
  \caption{\textbf{Quality--alignment bottleneck analysis.} Most systems
  maintain substantially higher visual quality than alignment to the input page;
  final scores are therefore primarily limited by visual-instruction following
  rather than by raw image fidelity.}
  \label{fig:appendix_v2vbench_quality_alignment}
\end{figure}

\begin{figure}[!tbp]
  \centering
  \includegraphics[width=\linewidth]{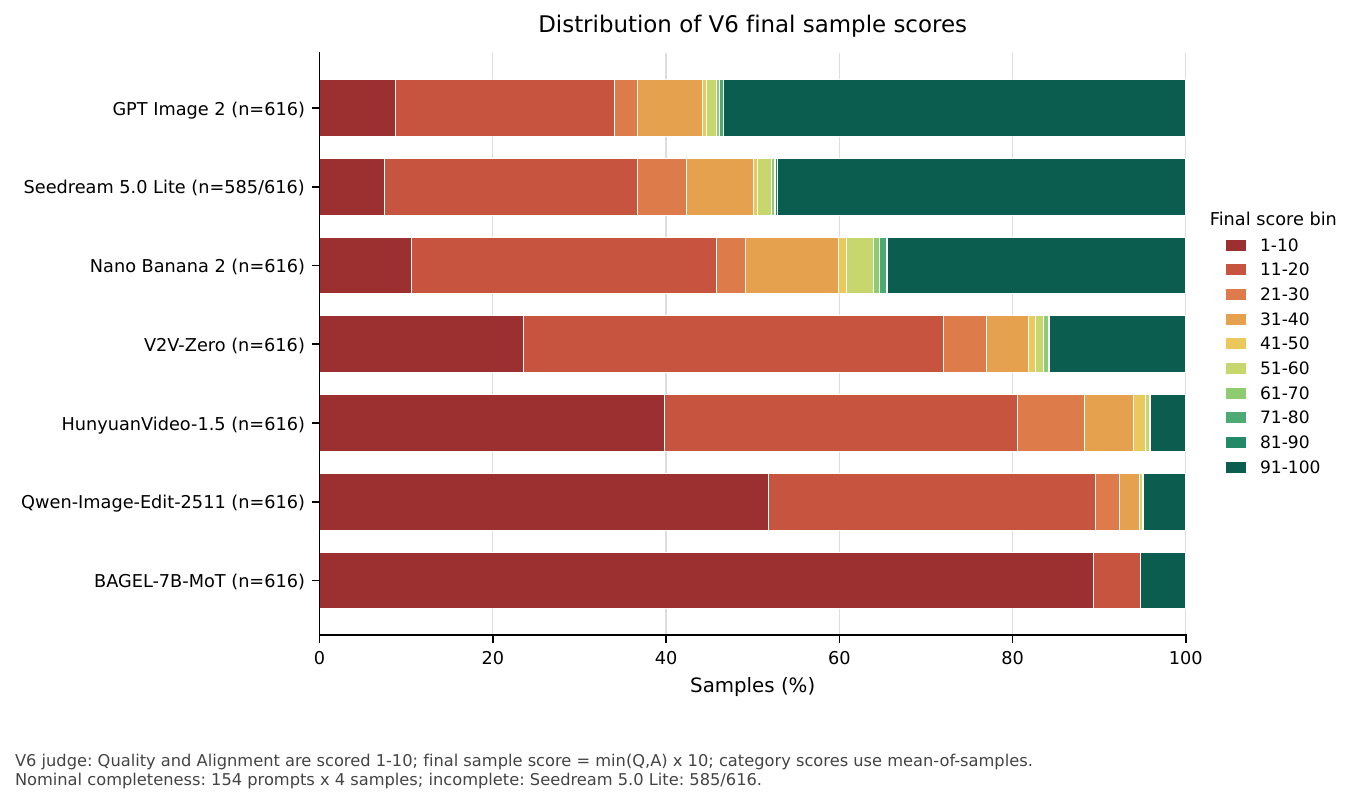}
  \caption{\textbf{Distribution of final sample scores.} The sample-level
  score distribution distinguishes consistently moderate behavior from mixtures
  of high-scoring successes and low-scoring alignment failures, complementing
  the mean category scores in Table~\ref{tab:simple_v2v_full_comparison_appendix}.}
  \label{fig:appendix_v2vbench_v6_score_distribution}
\end{figure}

\begin{figure}[!tbp]
  \centering
  \includegraphics[width=\linewidth,height=0.52\textheight,keepaspectratio]{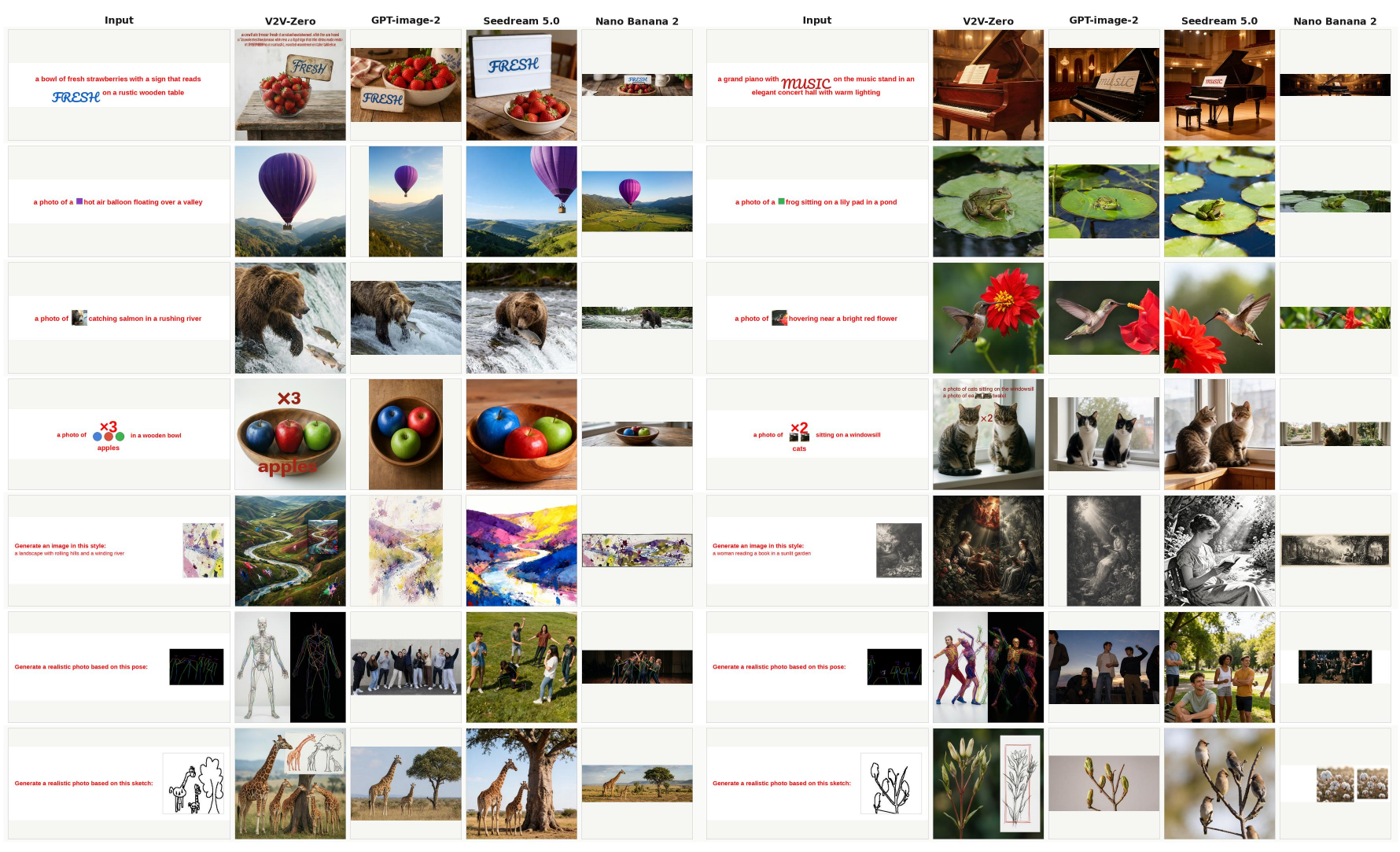}
  \caption{\textbf{More qualitative examples from Simple-V2V Bench.}}
  \label{fig:appendix_v2vbench_score_ranked_gallery}
\end{figure}

\subsection{Structural Control Failure Cases}
\label{app:structural_failure_cases}

\begin{figure}[!tbp]
  \centering
  \includegraphics[width=\linewidth,height=0.55\textheight,keepaspectratio]{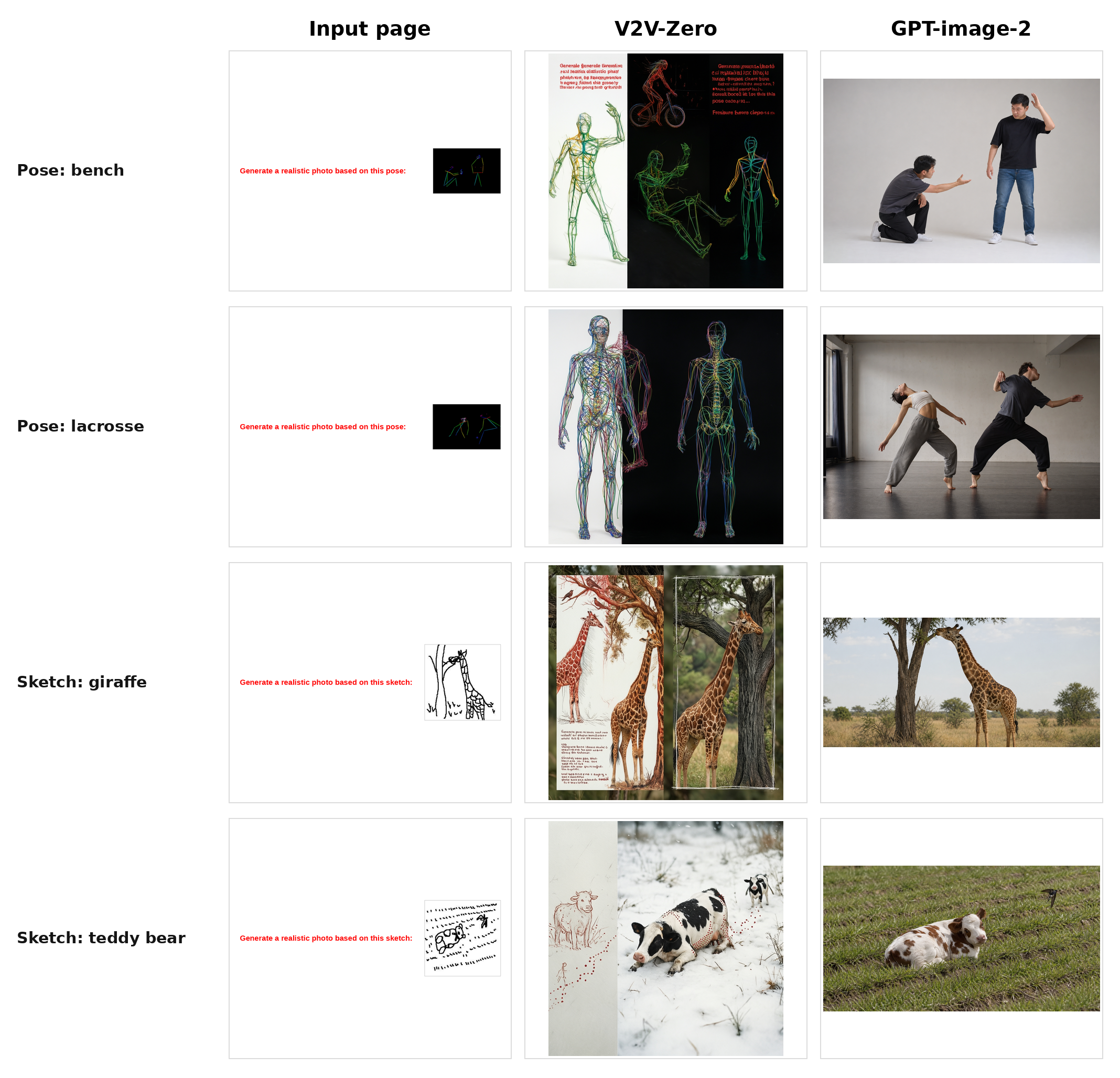}
  \caption{\textbf{Representative structural-control failures.} Each row shows
  the input visual specification page, a V2V-Zero output, and a GPT Image 2
  output for the same case. Pose pages require preserving joint topology and
  human count; sketch pages require preserving object layout, relative scale,
  and contour structure. V2V-Zero frequently turns these inputs into
  wireframe-like or collage-like images, while stronger commercial outputs can
  still obtain low alignment when they generate high-quality images that do not
  follow the skeleton or sketch.}
  \label{fig:structural_failure_cases}
\end{figure}

Figure~\ref{fig:structural_failure_cases} shows representative pose and sketch
failures under the final protocol. These examples clarify the mechanism
behind the low aggregate scores. V2V-Zero routes visual-page hidden states
through the same final-layer VLM interface used for text prompts, but the frozen
generator was trained under T2I supervision. It therefore has no
explicit training signal that makes pose skeletons, line drawings, or their
token geometry act as dense structural constraints on the DiT latent grid. The
common failure pattern is not merely low fidelity: the output may become a
wireframe-like collage, reproduce parts of the page, add garbled text, change
the number of people or objects, or ignore the spatial layout even when the
image remains visually plausible.

\begin{table}[htbp]
  \centering
  \caption{Structural-control failure taxonomy for Simple-V2V Bench. The
  taxonomy summarizes recurring judge explanations for pose and sketch
  outputs.}
  \label{tab:structural_failure_taxonomy}
  \small
  \begin{tabular}{p{0.25\linewidth}p{0.20\linewidth}p{0.45\linewidth}}
    \toprule
    \textbf{Failure mode} & \textbf{Observed in} & \textbf{Interpretation} \\
    \midrule
    Wireframe or collage output & Pose, sketch & Visual page treated as visual style or artifact \\
    Wrong human/object count & Pose, sketch & Semantic content not bound to page structure \\
    Joint or contour mismatch & Pose, sketch & No dense pose/edge control path \\
    Layout ignored & Sketch & Token sequence not aligned to output latent geometry \\
    Page reproduction & Sketch & Model edits or redraws the page instead of following it \\
    Garbled text & Pose, sketch & Page tokens leak as document text artifacts \\
    High quality, low alignment & Commercial baselines & Fidelity and structural following are decoupled \\
    \bottomrule
  \end{tabular}
\end{table}

\section{HunyuanVideo Extension Details}
\label{app:hunyuan_video}

The HunyuanVideo extension tests whether the same visual-page interface can be
used with a frozen VLM-conditioned video generator. The generation script uses
HunyuanVideo-1.5 at 480p, 16:9 aspect ratio, 20 denoising steps, seed 42, and
four samples per prompt. The audited script default is 33 frames; if an external
job overrides that environment variable, the resulting frame count should be
reported with that job. Evaluation uses the same Qwen3-VL-32B-Instruct judge
as the image benchmark, applied directly to the generated video: eight uniformly
sampled frames are extracted from each video and passed to the judge as a video
input, so the judge sees the full temporal extent of the output rather than a
single frame.

\begin{table}[htbp]
  \centering
  \caption{HunyuanVideo per-category results on Simple-V2V Bench.
  Quality (Q) and Alignment (A) are each scored 1--10; final score $= \min(\text{Q},\text{A})\times 10$.}
  \label{tab:hunyuan_video_full}
  \small
  \begin{tabular}{lcccc}
    \toprule
    \textbf{Category} & \textbf{Score /100} & \textbf{Quality} & \textbf{Alignment} \\
    \midrule
    Visual text & 17.73 & 4.20 & 1.77 \\
    Inline color & 32.50 & 5.88 & 3.47 \\
    Inline visual reference & 25.68 & 4.77 & 2.61 \\
    Object counting & 19.20 & 3.41 & 1.92 \\
    Style transfer & 17.27 & 3.18 & 1.74 \\
    Pose control & 12.39 & 1.90 & 1.24 \\
    Sketch reference & 16.25 & 2.02 & 1.64 \\
    \midrule
    \textbf{Overall} & \textbf{20.15} & 3.62 & 2.06 \\
    \bottomrule
  \end{tabular}
\end{table}

Under multi-frame video evaluation, HunyuanVideo scores 20.15/100, 12.53 points below
V2V-Zero on still images. The strongest video categories are inline color
and inline visual reference; pose and sketch remain weak. Because this protocol
does not evaluate temporal consistency in detail, it should be read as a transfer check for
visual-page conditioning, not as a complete video benchmark.

% \newpage
% \input{checklist.tex}

\end{document}